\documentclass{article}

\usepackage{microtype}
\usepackage{graphicx}
\usepackage{subcaption}
\usepackage{booktabs} 

\usepackage{hyperref}

\usepackage[preprint]{icml2026}

\usepackage{amssymb}
\usepackage{mathtools}
\usepackage{amsthm}
\usepackage{multirow}
\usepackage{tabularx}
\usepackage{threeparttable}
\usepackage{enumitem}
\usepackage{nicefrac}       
\usepackage{microtype}      
\usepackage{xcolor}         
\usepackage{setspace}
\usepackage{latexsym}
\usepackage{graphicx}
\usepackage{amsmath}
\usepackage{pifont}   
\usepackage{xcolor}   
\usepackage{amssymb} 
\usepackage{xcolor}
\usepackage{soul}
\usepackage{hyperref}
\usepackage{colortbl}
\usepackage{fontawesome}
\usepackage{pifont}
\usepackage[capitalize,noabbrev]{cleveref}

\theoremstyle{plain}

\theoremstyle{definition}

\theoremstyle{remark}

\definecolor{myblue}{RGB}{220,230,250}
\definecolor{bestcell}{HTML}{F2F2FE}
\usepackage[textsize=tiny]{todonotes}
\usepackage[most]{tcolorbox}

\icmltitlerunning{Repairing and Rewarding Tool-Use Trajectories for Tool-Integrated Reasoning}

\begin{document}

\twocolumn[
  \icmltitle{\textit{Guided by Trajectories:}
 Repairing and Rewarding Tool-Use Trajectories for Tool-Integrated Reasoning}



  \icmlsetsymbol{equal}{*}

  \begin{icmlauthorlist}
    \icmlauthor{Siyu Gong}{yyy,comp}
    \icmlauthor{Linan Yue*}{yyy,comp}
    \icmlauthor{Weibo Gao}{sch}
    \icmlauthor{Fangzhou Yao}{sch}
    \icmlauthor{Shimin Di}{yyy,comp}
    \icmlauthor{Lei Feng}{yyy,comp}
    \icmlauthor{Min-Ling Zhang}{yyy,comp}
  \end{icmlauthorlist}

  \icmlaffiliation{yyy}{School of Computer Science and Engineering, Southeast
University, Nanjing, China}
  \icmlaffiliation{comp}{Key Lab. of Computer Network
and Information Integration (Southeast University), MOE, China}
  \icmlaffiliation{sch}{State Key Laboratory of Cognitive Intelligence, University of Science and Technology of China, Hefei, China}

  \icmlcorrespondingauthor{Linan Yue}{lnyue@seu.edu}


  \vskip 0.3in
]



\printAffiliationsAndNotice{}  
\begin{abstract}
Tool-Integrated Reasoning (TIR) enables large language models (LLMs) to solve complex tasks by interacting with external tools, yet existing approaches depend on high-quality synthesized trajectories selected by scoring functions and sparse outcome-based rewards, providing limited and biased supervision for learning TIR.
 To address these challenges, we propose AutoTraj, a two-stage framework that automatically learns TIR by repairing and rewarding tool-use trajectories. Specifically, in the supervised fine-tuning (SFT) stage, AutoTraj generates multiple candidate tool-use trajectories for each query and evaluates them along multiple dimensions. High-quality trajectories are directly retained, while low-quality ones are repaired using an LLM (i.e., LLM-as-Repairer). The resulting repaired and high-quality trajectories form a synthetic SFT dataset, while each repaired trajectory paired with its original low-quality counterpart constitutes a dataset for trajectory preference modeling. In the reinforcement learning (RL) stage, based on the preference dataset, we train a trajectory-level reward model to assess the quality of reasoning paths and combine it with outcome and format rewards, thereby explicitly guiding the optimization toward reliable TIR behaviors.
 Experiments on real-world benchmarks demonstrate the effectiveness of AutoTraj in TIR.
 Code is available at \url{https://github.com/GSYY0607/AutoTraj_code}.
 \vspace{-0.2cm}
\end{abstract}

\section{Introduction}
In recent years, large language models (LLMs) have demonstrated remarkable capabilities in natural language processing and complex reasoning tasks, driving rapid advances across a wide range of artificial intelligence applications \cite{havrilla2024glore,minaee2024large,yue2025don,kumar2025llm,hurst2024gpt}.  However, in real-world reasoning scenarios, models often need to interact with external environments to coordinate multiple abilities to solve complex problems, such as information retrieval \cite{song2025r1} and code execution \cite{hosain2025xolver}. To this end, Tool-Integrated Reasoning (TIR) \cite{paranjape2023art,gou2023tora,wei2025autotir,baek2025researchagent} has been proposed to enable models to invoke external tools during reasoning, thereby extending their problem-solving capabilities and improving practical performance.

\begin{table}[t]
\centering
    \vspace{-0.1cm}
\caption{Differences between existing two-stage TIR paradigms and AutoTraj. Compared with previous methods, 
AutoTraj repairs low-quality TIR trajectories in Stage 1 for SFT and introduces trajectory-aware reinforcement learning in Stage 2 by leveraging automatically synthesized positive and negative trajectory pairs, enabling more scalable TIR.
}

  \setlength{\abovecaptionskip}{0.1cm}
    \setlength{\belowcaptionskip}{0.2cm}
      \renewcommand\arraystretch{1.1}
\setlength{\tabcolsep}{0.1mm}{
    \scalebox{0.72}{
\begin{tabular}{lcc}
\toprule
Property & Existing Two-stage TIR & AutoTraj (ours) \\
\midrule
\multicolumn{3}{c}{{Stage 1: Supervised Fine-tuning (SFT)}} \\
Synthesized  trajectories& \textcolor[HTML]{27CC30}{\ding{51}}  & \textcolor[HTML]{27CC30}{\ding{51}}  \\
 High-quality TIR trajectories & Retain  & Retain \\
 Low-quality TIR trajectories & Filter  & Repair \\

\midrule
\multicolumn{3}{c}{{Stage 2: Reinforcement Learning (RL)}} \\
Reward target & Answer&Answer+Trajectory\\
Format Reward &\textcolor[HTML]{27CC30}{\ding{51}}  & \textcolor[HTML]{27CC30}{\ding{51}}  \\
Outcome Reward &\textcolor[HTML]{27CC30}{\ding{51}} & \textcolor[HTML]{27CC30}{\ding{51}}  \\
Trajectory Reward &\textcolor{red}{\ding{55}}  & \textcolor[HTML]{27CC30}{\ding{51}}  \\
Trajectory RM Training&$\varnothing$& Synthesized trajectory pairs\\

\bottomrule
\end{tabular}
  }
    }   
    \label{table-different}
    \vspace{-0.7cm}
\end{table}

With the recent progress of large reasoning models (e.g., DeepSeek R1 \cite{guo2025deepseek}) in long-chain reasoning and complex decision-making, researchers have begun to transfer these reasoning paradigms to the TIR setting. Specifically, existing approaches typically adopt a two-stage training paradigm to enhance TIR abilities \cite{dong2025tool,dong2025agentic,feng2025retool}. As shown in Table \ref{table-different}, \textit{In Stage 1}, models are initialized through supervised fine-tuning (SFT) on tool-use trajectories to achieve a cold start. \textit{In Stage 2}, the model is further optimized using reinforcement learning (RL) with outcome-based rewards to improve its ability to leverage tools for solving complex multi-step reasoning tasks. While this paradigm improves TIR capabilities to some extent, it still faces substantial challenges in practice.

Specifically, in the first, \textbf{\textit{during the SFT stage}}, acquiring high-quality expert tool trajectories is costly and difficult to scale \cite{yao2022react,schick2023toolformer,acikgoz2025can}. Therefore, as shown in Table~\ref{table-different}, most existing approaches \cite{dong2025tool,feng2025retool} adopt automatically synthesized trajectories as a practical alternative to human-labeled ones, typically following a pipeline that generates a large pool of candidates and then selects higher-quality samples via heuristic scoring. However, this  ``synthesize-then-filter'' paradigm tends to reduce data diversity. Specifically, the synthesis model is usually proficient in only certain reasoning patterns, producing high-quality trajectories in familiar cases but low-quality ones in challenging scenarios. Simply filtering these imperfect samples may lead to a biased training distribution, limiting robustness and generalization. Moreover, low-quality trajectories are not merely noise but informative negative examples, filtering them out wastes valuable supervision signals for distinguishing correct from incorrect reasoning paths.
Second, \textbf{\textit{during the RL stage}}, training strategies that rely solely on outcome-based rewards suffer from reward sparsity \cite{feng2025retool,li2025torl,deng2025atom,lightman2023let}.  Such rewards focus exclusively on the correctness of the final answer while ignoring the quality of the underlying TIR path. As a result, trajectories with flawed reasoning but accidentally correct outcomes may be rewarded, whereas trajectories with sound reasoning but minor final errors may be penalized, hindering the learning of stable and reliable tool-integrated reasoning behaviors.

In this paper, we propose \textbf{AutoTraj}, a novel framework that learns tool-integrated reasoning from \underline{\textbf{Auto}}matically repairing and rewarding tool-use \underline{\textbf{Traj}}ectories. AutoTraj also follows a two-stage training paradigm. \textbf{\textit{In the SFT stage}}, our goal is to synthesize high-quality tool-use trajectories. Specifically, given an input query, we first use a base model equipped with tool-calling capabilities to generate multiple candidate reasoning trajectories that involve tool usage. These trajectories are then jointly evaluated along multiple dimensions, including answer correctness, reasoning confidence, trajectory length, and text repetition, and categorized into high-quality and low-quality groups. Instead of simply discarding low-quality trajectories, we introduce a trajectory repair mechanism that leverages an LLM to repair and complete their TIR trajectories (LLM-as-Repairer).
The repaired trajectories, together with the original high-quality trajectories, form a high-quality dataset $\mathcal{D}_{\text{sft}}$ for SFT, which preserves diverse reasoning patterns from originally low-quality samples. Meanwhile, we further construct positive–negative trajectory pairs by pairing each repaired trajectory with its corresponding original low-quality trajectory, resulting in an auxiliary dataset $\mathcal{D}_{\text{self}}$ that provides fine-grained supervision for subsequent reward modeling. At this stage, SFT on $\mathcal{D}_{\text{sft}}$ improves the model’s basic TIR capabilities.
\textbf{\textit{In the RL stage}}, to mitigate reward sparsity, we first train a Trajectory Reward Model (Trajectory RM) on $\mathcal{D}_{\text{self}}$ to distinguish high-quality and low-quality tool-use trajectories. We then optimize the model using Group Relative Policy Optimization (GRPO) \cite{shao2024deepseekmath}. In addition to standard format and outcome rewards, we incorporate reward scores from the Trajectory RM to evaluate the generated reasoning trajectories. This design enables the optimization process to explicitly account for reasoning path quality,  alleviating reward sparsity and guiding the model toward more robust and reliable TIR strategies. Table \ref{table-different} summarizes the differences between the existing two-stage TIR methods and our proposed AutoTraj. 

Our main contributions are summarized as follows:
\begin{itemize}[leftmargin=*]
 \item We propose AutoTraj, a two-stage training framework for tool-integrated reasoning that reuses low-quality trajectories via automatic repair and incorporates trajectory-level rewards into RL, thereby providing more informative supervision and mitigating reward sparsity.
 \item  We construct two complementary datasets: a high-quality tool-use trajectory dataset for SFT, and a trajectory pair dataset consisting of positive–negative trajectory pairs for training a trajectory reward model. The learned reward model can also provide fine-grained trajectory-level rewards for training other TIR models.
 \item Extensive experiments on diverse benchmark datasets, covering mathematical reasoning and knowledge-intensive reasoning, show that the proposed framework consistently achieves substantial improvements in both tool-use capability and reasoning quality.
\end{itemize}

\section{Related Work}
\subsection{Tool-Integrated Reasoning}
\label{section 2.1}
Tool-Integrated Reasoning (TIR) \cite{schick2023toolformer,paranjape2023art,li2025deepagent,inoue2025drugagent} has recently emerged as a prominent paradigm for enhancing the complex reasoning capabilities of LLMs by enabling them to autonomously invoke external tools. Existing studies on TIR can be broadly categorized into three lines of research.
The first line of work relies on supervised fine-tuning (SFT-based TIR) over manually annotated reasoning trajectories with explicit tool calls, such as \cite{chen2022program, gou2023tora, li2023chain, schick2023toolformer}, which can effectively teach models specific tool-use patterns but suffer from high annotation costs and limited scalability to open-world scenarios. 
The second line, RL-based TIR, adopts reinforcement learning with verifiable rewards, where the correctness of the final answer or feedback from an external verifier serves as the reward signal, as in ToRL \cite{li2025torl}, ToolRL \cite{qian2025toolrl}, ReSearch \cite{chen2025learning}, Search-R1 \cite{jin2025search}, ARTIST \cite{singh2025agentic}, and AutoTIR \cite{wei2025autotir}. These outcome-only rewards are extremely sparse, offering little guidance for intermediate reasoning steps and resulting in unstable tool-use trajectories.
Finally, the third stream combines SFT with RL, initializing models with synthesized trajectories and then further optimizing them via RL, as in Tool-Star \cite{dong2025tool} and ReTool \cite{feng2025retool}. Although this hybrid paradigm is effective in practice, it still inherits the dependence on high-quality synthesized trajectories selected by scoring functions and the coarse-grained nature of the outcome-only reward. In contrast, our proposed AutoTraj repairs low-quality synthesized trajectories instead of simply filtering them, and further introduces a trajectory-level reward to provide fine-grained supervision, alleviate reward sparsity, and guide the model toward more reliable TIR.

\subsection{Reinforcement Learning with Verifiable Rewards}
The core idea of Reinforcement Learning with Verifiable Rewards (RLVR) is to incorporate rule-based reward signals that can be automatically verified, such as answer correctness and format validity, into the RL framework. In reasoning tasks, RL has been shown to be an effective paradigm for enhancing the multi-step reasoning capabilities of LLMs. Representative works such as DeepSeek-R1 \cite{guo2025deepseek} and QwQ \cite{team2024qwq} demonstrate that directly optimizing policies with respect to outcome correctness can significantly improve overall reasoning performance. Building on this idea, RLVR has been further extended to tool-integrated reasoning scenarios to learn more stable and effective tool invocation and reasoning strategies (as reviewed in Section~\ref{section 2.1}). However, existing RLVR methods mostly rely on outcome-only verifiable rewards, which results in highly sparse training signals and makes it difficult to effectively constrain intermediate reasoning processes and tool-use trajectories. Therefore, recent studies have begun to introduce process-level or trajectory-level supervision \cite{3ZWCCELKKKF025,wang2025steca,deng2025atom,wang2025answer} to encourage models to generate more reliable intermediate reasoning behaviors. However, trajectory-level reward design for TIR remains under-explored and is still insufficient to provide effective supervision over tool-use and multi-step reasoning processes.

\section{Problem Formulation}
Given a query $q$, its ground-truth answer $y$, and an environment $\varphi(\cdot)$ that provides access to a set of external tools, our goal is to train a TIR model $\pi_\theta(\cdot)$. The model is expected not only to predict the final answer, but also to generate a corresponding tool-integrated reasoning trajectory $\tau$ that explicitly captures the reasoning and tool-use process.
Formally, a reasoning trajectory $\tau$ is defined as a sequence of $K$ reasoning steps:
\begin{eqnarray}
  \tau = {(r_k, t_k, e_k)}_{k=1}^{K},
\end{eqnarray}
where at step $k$, $r_k$ denotes the natural language reasoning generated by the model, $t_k$ denotes the invoked tool action, and $e_k$ represents the execution result returned by the environment $\varphi$ after applying $t_k$.

The trajectory is generated in an iterative manner. At each step $k$, the model conditions on the input query and the previously generated partial trajectory $ \tau_{k-1}$, and produces the next reasoning and tool invocation:
\begin{eqnarray}
(r_k, t_k) = \pi_\theta(q \oplus \tau_{k-1}), \quad
e_k = \varphi(t_k),
\end{eqnarray}
where  $\oplus$ represents concatenation.
This iterative generation loop continues until the model produces a final answer or reaches a predefined maximum context length.

In this paper, we employ a search engine and code execution as the tools, which serve as representative tools to achieve AutoTraj.

\section{Framework of AutoTraj}
In this section, as shown in Figure \ref{model}, we present the details of AutoTraj, consisting of two stages:

$\bullet$ Supervised Fine-Tuning via Synthesized Trajectory (SFT Stage): In this stage, we automatically generate and filter TIR trajectories and repair low-quality ones. The resulting supervision dataset $\mathcal{D}_{\text{sft}}$ is composed of both the original high-quality trajectories and the repaired ones.
This dataset is then used for supervised cold start. Meanwhile, we create the trajectory pairs by pairing repaired trajectories with their corresponding original low-quality trajectories, forming a dataset $\mathcal{D}_{\text{self}}$ for trajectory preference modeling.

$\bullet$ Trajectory-Aware Reinforcement Learning (RL stage): In this stage, we first train a trajectory reward model based on $\mathcal{D}_{\text{self}}$ that can be used to evaluate the quality of reasoning trajectories. We then optimize the TIR model using the RL method, combining the trajectory reward with standard outcome and format rewards, thereby explicitly guiding the model to generate both correct and high-quality tool-integrated reasoning paths.

\subsection{Supervised Fine-Tuning via Synthesized Trajectory}
\begin{figure*}[t]
  \centering 
  \includegraphics[width = 13.3cm]{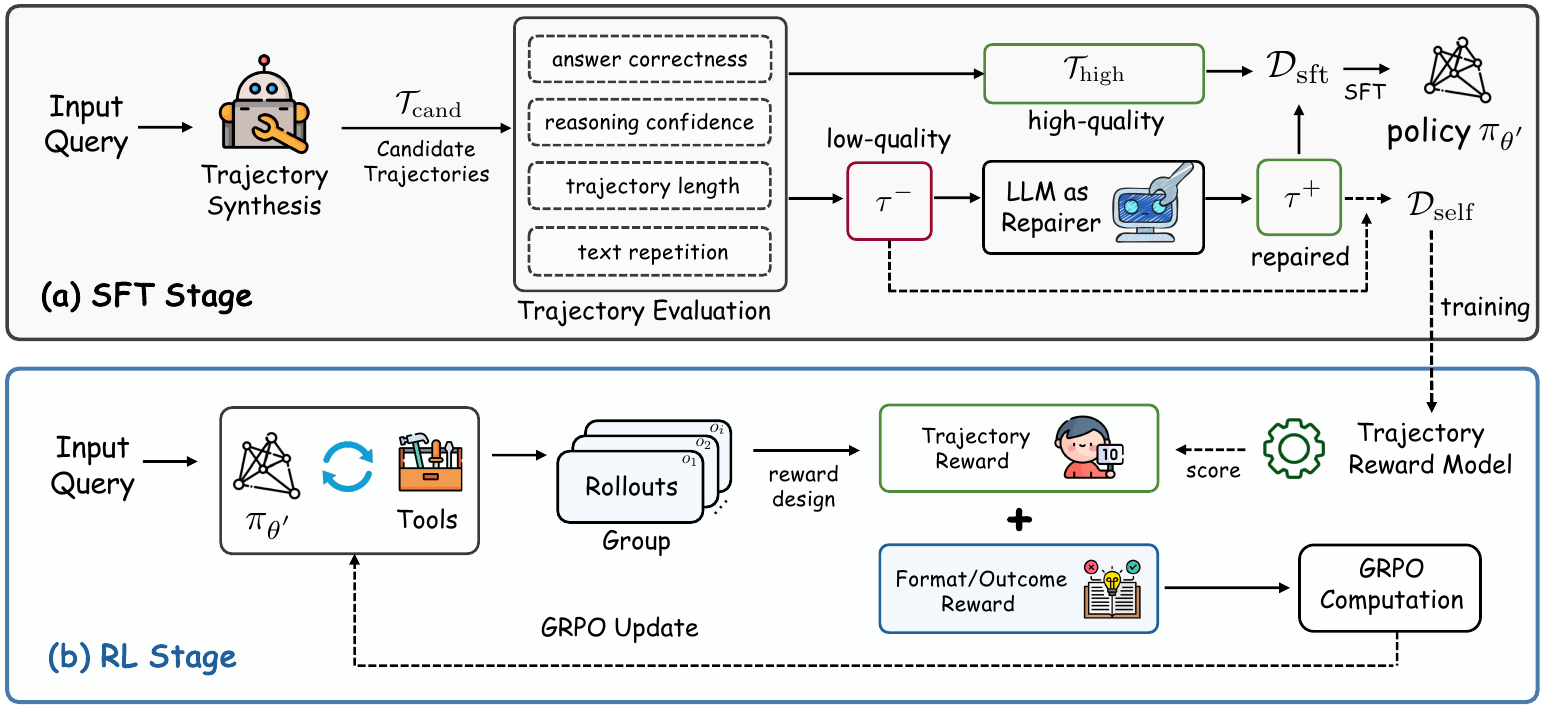}
 \caption{
  Architecture of AutoTraj. (a) SFT Stage: AutoTraj first synthesizes, evaluates, and repairs tool-use trajectories to construct a high-quality SFT dataset and positive–negative trajectory pairs. (b) RL Stage: A trajectory reward model is then trained on these pairs and integrated with format and outcome rewards under GRPO, providing trajectory-level supervision.
 }
 \label{model}
\end{figure*}

In this subsection, as shown in Figure \ref{model}(a), we first describe how to automatically construct tool-integrated trajectories. We then introduce our multi-dimensional metrics for evaluating these trajectories and detail how low-quality trajectories are repaired. Finally, the combination of repaired trajectories and the original high-quality trajectories is used for cold-start supervised fine-tuning.

\subsubsection{Candidate TIR Trajectory Synthesis}
\label{section-4-1-1}
To synthesize an initial set of candidate TIR data, we sample 10,000 question–answer pairs ${(q, y)}$ from the Tool-Star dataset~\cite{dong2025tool}, and discard all annotated tool-use trajectories to ensure independence.
Then, we denote the pretrained trajectory generator as $f_\psi(\cdot)$, which is  used to generate $n$ candidate trajectories based on the given query~$q$:
\begin{eqnarray}
 \mathcal{T}_{\text{cand}} = \{\tilde{\tau}_i\}_{i=1}^n  \sim f_\psi(\tilde{\tau} \mid q),
\end{eqnarray}
where each $\tilde{\tau}_i = \{ (\tilde{r}_k, \tilde{t}_k, \tilde{e}_k) \}_{k=1}^{K_i}$ represents a candidate sequence of \textit{reasoning-action-observation} tuples.

To enforce a structured and machine-parsable format, we design a constrained prompting protocol (see Appendix~\ref{prompt1}). Specifically, each trajectory starts with a \verb|<think>| block corresponding to the reasoning variable $\tilde{r}_k$. At each step, the model decides whether to invoke a tool $\tilde{t}_k \in \{\texttt{code}, \texttt{search}\}$, and the resulting observation $\tilde{e}_k = \varphi(\tilde{t}_k)$ is inserted between \verb|<result>| and \verb|</result>| tags, forming the TIR trajectoriy $(\tilde{r}_k, \tilde{t}_k, \tilde{e}_k)$. This iterative generation continues until a terminal answer $\hat{y}$ is produced or a predefined maximum context length is reached. Finally, the answer is enclosed in \verb|\boxed{}|. In practice, to prevent trajectories from becoming excessively long, we limit each candidate trajectory to at most three tool invocations.

\subsubsection{Multi-Dimensional Trajectory Evaluation}
\label{section-4-1-2}
To assess the overall quality of TIR trajectories, we design a multi-dimensional evaluation that jointly considers answer correctness, reasoning confidence, trajectory length, and text repetition\footnote{We select these metrics based on observations from our preliminary experiments, which show that low-quality trajectories often exhibit low reasoning confidence, suboptimal lengths, or excessive repetition.}. Based on these multi-dimensional criteria, each candidate trajectory $\tilde{\tau}_i \in \mathcal{T}_{\text{cand}}$ is assigned a composite quality score and subsequently classified into different quality levels for downstream training.

Specifically, answer correctness is used for initial screening. Trajectories producing an incorrect answer $\hat{y}(\tilde{\tau}_i) \neq y$ are immediately assigned to the low-quality set: $\tilde{\tau}_i \in \mathcal{T}_\text{low}^{\text{wrong}}.$
Only trajectories with correct answers will be proceed to further evaluation.
Then, for each correct-answer trajectory, a multi-dimensional quality score is defined as:
\begin{eqnarray}
  S(\tilde{\tau}_i) = \lambda_{1} \, S_\text{conf}(\tilde{\tau}_i) + \lambda_{2} \, S_\text{len}(\tilde{\tau}_i) + \lambda_{3} \, S_\text{rep}(\tilde{\tau}_i),
  \label{multi}
\end{eqnarray}
where $S_\text{conf}(\cdot), S_\text{len}(\cdot), S_\text{rep}(\cdot)$ denote the confidence, length, and text repetition score, respectively and $\lambda_{1}+\lambda_{2}+\lambda_{3}=1$.
Next, we describe how these scores are computed:

$\bullet$ For the confidence score $S_{\text{conf}}(\tilde{\tau}_i)$, we aim to assess whether a trajectory exhibits hesitation or redundant self-corrections. To this end, we define a binary confidence measure: 
if the TIR trajectory contains explicit lexical cues of self-doubt or uncertainty, such as \textit{(maybe, unsure, guess, seems, perhaps, probably, recheck, related terms)}, its confidence score $S_{\text{conf}}(\tilde{\tau}_i)$ is set to 0. Otherwise, it is set to~1.

$\bullet$  For the length score $S_{\text{len}}(\tilde{\tau}_i)$, we aim to quantify how much a candidate trajectory deviates from the ideal TIR length for a given query $q$: 
$S_\text{len}(\tilde{\tau}_i) = \exp(-(\text{len}(\tilde{\tau}_i) - l_{q})^2 / (2\sigma^2))$,
where, in practice, the ideal length $l_q$ is defined as the shortest length among the trajectories in $\mathcal{T}_{\text{cand}}$ that produce the correct answer.
This metric is designed to assign lower scores to trajectories that are excessively long.

$\bullet$ For the repetition score, we aim to assess the degree of repetition in local text within a trajectory. Let $\text{Rep}_N(\tilde{\tau}_i)$ denote the $N$-gram repetition rate. We define the repetition score as:
$
S_{\text{rep}}(\tilde{\tau}_i) = 1 - \text{Rep}_N(\tilde{\tau}_i),
$
which penalizes low-information or meaningless repeated content.

Finally, we set a quality threshold $\theta_\text{qual}$ such that a trajectory $\tilde{\tau}_i$ is considered high-quality if $S(\tilde{\tau}_i)$ exceeds  $\theta_\text{qual}$, and low-quality otherwise\footnote{For detailed implementation, see Appendix \ref{training-setting}.}.
Ultimately, all candidate trajectories are divided into three disjoint sets: $\mathcal{T}_\text{high}$, $\mathcal{T}_\text{low}^{\text{correct}}$, and  $\mathcal{T}_\text{low}^{\text{wrong}}$.

\subsubsection{LLM-based Trajectory Repair}
\label{section-4-1-3}
To repair low-quality trajectories, we treat the LLM as a trajectory repairer (i.e., LLM-as-Repairer) to automatically refine low-quality TIR trajectories. Specifically, given a query $q$ and its corresponding low-quality trajectory $\tau^{-} \in \mathcal{T}_{\text{low}}^{\text{correct}}$ or $\tau^{-} \in \mathcal{T}_{\text{low}}^{\text{wrong}}$, we feed both into the repairer and prompt it to regenerate a candidate repaired trajectory $\tau^{+}$. We then evaluate the quality of $\tau^{+}$ using our scoring function. If its score exceeds the predefined high-quality threshold, the repair is regarded as successful. The repaired trajectory is added to the high-quality trajectory set $\mathcal{T}_{\text{high}}$, and meanwhile, the pair $(\tau^{-}, \tau^{+})$ is stored as a contrastive sample in the self-supervised dataset $\mathcal{D}_{\text{self}}$.

For different types of low-quality trajectories ($\mathcal{T}_{\text{low}}^{\text{correct}}$ and $\mathcal{T}_{\text{low}}^{\text{wrong}}$), we adopt prompt designs with different emphases during the repair process (see Appendix~\ref{prompt2}), so as to respectively focus on reasoning refinement and error diagnosis.

Finally, based on the above procedure (Sections~\ref{section-4-1-2} and~\ref{section-4-1-3}), we construct two datasets. Among them, the supervised fine-tuning dataset is defined as:
\begin{equation}
\mathcal{D}_{\text{sft}} = \{ (q_i, y_i, \tau_{i} \}_{i=1}^{N},
\end{equation}
where $(q_i, y_i)$ denotes the $i$-th query--answer pair, and
$
\tau_{i} \in \mathcal{T}_{\text{high}}(q_i, y_i) = \{ \tau_{i,1}^{+}, \ldots, \tau_{i,m_i}^{+} \}
$
is the set of high-quality trajectories (including both originally high-quality ones and successfully repaired ones) associated with $(q_i, y_i)$. 

The self-supervised dataset is defined as:
\begin{equation}
\mathcal{D}_{\text{self}} = \{ (q_j,\tau_j^{-}, \tau_j^{+}) \}_{j=1}^{M},
\end{equation}
where $\tau_j^{-}$ is an original low-quality trajectory and $\tau_j^{+}$ is its successfully repaired counterpart that meets the high-quality threshold. Each element thus forms a low--high trajectory pair for subsequent reward model training.

\subsubsection{Cold-Start Supervised Fine-tuning}
After constructing the SFT dataset $\mathcal{D}_{\text{sft}}$, we perform cold-start supervised fine-tuning to equip AutoTraj with an initial capability for solving problems using tools. Specifically, given a sample $(q_i, y_i, \tau_i) \in \mathcal{D}_{\text{sft}}$, we optimize the policy model $\pi_\theta$ to $\pi_{\theta^{'}}$ by minimizing the negative log-likelihood: 
\begin{equation}
\mathcal{L}_{\text{SFT}} = - \mathbb{E}_{(q_i,y_i, \tau_i) \sim \mathcal{D}_{\text{sft}}}
\left[\log P_\theta((\tau_i,y_i) \mid q_i) \right].
\end{equation}

\subsection{Trajectory-Aware Reinforcement Learning}
After SFT, in this subsection,  as shown in Figure \ref{model}(b), we first train a trajectory reward model. Using this model, we design a comprehensive reward function that integrates format, outcome, and trajectory rewards for RL training.
\subsubsection{Trajectory Reward Model}
\label{section-4-3-1}
To evaluate trajectory quality during RL, a trajectory-level reward model $r_\phi(\cdot)$ is required. Since such a model is generally unavailable, we train using our self-supervised dataset $\mathcal{D}_{\text{self}}$.
Specifically, the reward model is optimized to assign higher scores to high-quality trajectories than to low-quality trajectories for the same query. To this end, we adopt a pairwise ranking objective:
\begin{equation}
\mathcal{L}_{\text{RM}} = - \mathbb{E}_{(q, \tau^{+}, \tau^{-}) \sim \mathcal{D}_{\text{self}}} \left[ \log \sigma\big(r_\phi(q, \tau^{+}) - r_\phi(q, \tau^{-})\big) \right],
\end{equation}
where $\sigma(\cdot)$ denotes the sigmoid function, and the input query $q$ is included in the prompt fed to the model. 

Through this training, the resulting trajectory reward model can effectively distinguish high- and low-quality trajectories and provide fine-grained trajectory-level scores. Compared with sparse answer-level rewards, it delivers denser and more stable supervision signals, which are then incorporated into the RL stage to guide the optimization of TIR behaviors.

\subsubsection{Reward Design}
To provide stable and fine-grained supervision for TIR, we design a hierarchical reward function composed of three components: a \emph{format reward}, an \emph{outcome reward}, and a \emph{trajectory reward}.

\textbf{Format Reward.}
We first introduce a binary format reward to strictly enforce the predefined output structure.
Given a model-generated response $(\tau_i, \hat{y}_i)$, the format reward $R_{\text{fmt}}$ is defined as:
\begin{equation}
R_{\text{fmt}} =
\begin{cases}
1,  & \text{if the output format is valid}, \\
-1, & \text{otherwise}.
\end{cases}
\end{equation}
The detailed format constraints have been described in Section~\ref{section-4-1-1}.

\textbf{Outcome Reward.}
The outcome reward is computed only when the format is valid ($R_{\text{fmt}} = 1$).
Specifically, we measure the token-level F1 score \cite{dong2025tool} between the predicted final answer $\hat{y}_i$ and the ground-truth answer~$y_i$:
\begin{equation}
R_{\text{ans}} = \text{F1}(y_i, \hat{y}_i).
\end{equation}

\textbf{Trajectory Reward.}
When the outcome reward is positive ($R_{\text{ans}} > 0$), we further introduce a trajectory-level reward to assess the overall quality of the reasoning process.
Concretely, the reasoning trajectory $\tau_i$ is evaluated by a trained trajectory reward model $r_\phi(\cdot)$, which outputs a scalar score:
\begin{equation}
R_{\text{traj}} =
\begin{cases}
r_\phi(q, \tau_i), & \text{if } r_\phi(q, \tau_i) > 0, \\
0, & \text{otherwise}.
\end{cases}
\end{equation}
where the trajectory reward is activated only for correct predictions, thereby preventing erroneous reasoning paths from receiving positive reinforcement.

\textbf{Final Reward.}
We combine the format, outcome, and trajectory rewards into a unified final reward:
\begin{equation}
R =
\begin{cases}
-1, & \text{if } R_{\text{fmt}} = -1, \\
0, & \text{if } R_{\text{fmt}} = 1 \text{ and } R_{\text{ans}} = 0, \\
R_{\text{ans}} + \alpha R_{\text{traj}}, & \text{if } R_{\text{fmt}} = 1 \text{ and } R_{\text{ans}} > 0,
\end{cases}
\end{equation}
where $\alpha$ is a weight that controls the contribution of the trajectory-level reward. In practice, to further stabilize RL, we impose an upper bound $R_{\max}$ on the final reward, so that all rewards are clipped into the range $[-1, R_{\max}]$.


\subsubsection{RL Training with GRPO}
\label{section-4-3-3}
To optimize the policy SFT model $\pi_{\theta^{'}}$, we adopt Group Relative Policy Optimization (GRPO) \cite{shao2024deepseekmath} as our RL algorithm, which estimates the baseline using a group of rollouts. Given a group of $G$ rollouts $\{o_i\}_{i=1}^G$ sampled from the current policy ${\pi}_\theta^{\text{old}}$ for query $q$, the GRPO objective is:
\begin{equation}
\tiny
\begin{aligned}
\mathcal{L}_{\text{GRPO}} =
& \mathbb{E}_{q, \{o_i\} \sim {\pi}_\theta^{\text{old}}} \Bigg[ 
 \frac{1}{G} \sum_{i=1}^{G} \frac{1}{|o_i|} \sum_{t=1}^{|o_i|} 
\min \Bigg(
\frac{\pi_{\theta^{'}}(o_{i,t}|q, o_{i,<t})}{{\pi}_\theta^{\text{old}}(o_{i,t}|q, o_{i,<t})} 
A_{i,t}^{\text{clip}}, \\
& \quad \text{clip}\Big(
\frac{\pi_{\theta^{'}}(o_{i,t}|q, o_{i,<t})}{{\pi}_\theta^{\text{old}}(o_{i,t}|q, o_{i,<t})}, 1-\epsilon, 1+\epsilon
\Big) A_{i,t} 
\Bigg) - \beta D_\mathrm{KL}[\pi_{\theta^{'}} || {\pi}_{\text{ref}}] 
\Bigg],
\end{aligned}
\end{equation}
where $\epsilon$ and $\beta$ are hyperparameters, $A_{i,t}$ is the normalized advantage of the $t$-th step in the $i$-th rollout, and $o_{i,<t}$ denotes the partial trajectory before step $t$. 
By optimizing this objective, we obtain our final AutoTraj that balances final answer correctness with high-quality reasoning behavior.

 \begin{table*}[t]
\centering
\caption{Performance comparison on mathematical and knowledge-intensive reasoning benchmarks. The top two results are highlighted in \textbf{bold} and \underline{underlined}. All methods adopt Qwen2.5-7B as the backbone model and are categorized into three groups: SFT-only, RL-only, and SFT–RL TIR methods. Among SFT–RL TIR methods, AutoTraj and Tool-Star are trained with the same RL dataset, while their SFT datasets differ substantially: Tool-Star uses 54K trajectories for SFT, whereas AutoTraj relies on only about 13K trajectories.}
  \renewcommand\arraystretch{1.25}
\setlength{\tabcolsep}{3.3mm}{
    \scalebox{0.81}{
    \setlength{\aboverulesep}{0.ex}
\setlength{\belowrulesep}{0.ex}
\begin{tabular}{lccccccccccc}
\toprule
\hline
\multirow{2.2}{*}{\textbf{Method}} 
& \multicolumn{5}{c}{\textbf{Mathematical Reasoning}} 
& \multicolumn{4}{c}{\textbf{Knowledge-Intensive Reasoning}} 
& \multirow{2.2}{*}{\textbf{Avg.}} \\
\cmidrule(lr){2-6} \cmidrule(lr){7-10} \addlinespace[-1.1em]\\
 & AIME24 & AIME25 & AMC23 & GSM8K & MATH & HQA & 2Wiki & MuSiQue & HLE &  \\
\midrule
Qwen2.5-7B-Instruct             & 6.67 & 0.00 & 27.50 & 46.78 & 57.30 & 24.50 & 20.50 & 10.50 & 5.60 & 22.15 \\
\midrule
\multicolumn{10}{l}{\textit{\textbf{SFT-only TIR Methods}}} \\
Tool-Star-SFT                   & 10.00 & \textbf{10.00} & 35.00 & 72.78 & 68.20 & 26.50 & 21.50 & 11.00 & 4.50 & 28.83\\
\midrule
\multicolumn{10}{l}{\textit{\textbf{RL-only TIR Methods}}} \\
ReSearch                        & 0.00 & 3.33 & 7.50 & 44.88 & 49.00 & \underline{28.50} & 26.50 & 14.00 & \underline{6.40} & 20.01 \\
R1-Searcher                     & 3.33 & 0.00 & 37.50 & 72.33 & 67.60 & 19.50 & 17.00 & 8.00 & 3.90 & 25.46 \\
ToRL                            & \textbf{23.33} & 3.33 & \underline{45.00} & 75.28 & 68.50 & 6.50 & 14.00 & 2.00 & 4.20 & 26.90 \\
AutoTIR                         & 6.67 & \underline{6.67} & 35.00 & 62.40 & 59.00 & \textbf{30.50} & 25.00 & \textbf{17.00} & 5.70 & 27.55 \\
\midrule
\multicolumn{10}{l}{\textit{\textbf{SFT-RL TIR Methods}}} \\
Vanilla SFT-RL TIR              & 3.33  & 3.33 & 42.50 & 78.85 & 69.00 & 28.00 & \underline{28.50} & 14.00 & 5.20 & 30.30 \\
Tool-Star                       & \underline{16.67} & \underline{6.67} & \underline{45.00} & \underline{79.91} & \textbf{69.50} & 27.00 & \textbf{29.50} & 9.00 & 4.60 & \underline{31.98} \\
\rowcolor{bestcell!90} AutoTraj & \textbf{23.33} & \textbf{10.00} & \textbf{47.50} & \textbf{79.98} & \underline{69.10} & \underline{28.50} & \textbf{29.50} & \underline{14.50} & \textbf{7.00} & \textbf{34.38} \\
\bottomrule
\hline
\end{tabular}
    }
    }  
\vspace{-0.1cm}
\label{tab:accuracy}
\end{table*}

\section{Experiments}
In this section, to demonstrate the effectiveness of AutoTraj, we design experiments to address the following research questions:
\begin{itemize}[leftmargin=*]
  \item \textbf{RQ1:}  Does AutoTraj improve TIR performance compared with baseline methods?

  \item \textbf{RQ2:} How do the individual components of AutoTraj contribute to overall performance, such as the contribution of the repair trajectory step?

  \item \textbf{RQ3:} How does the reasoning efficiency of AutoTraj compare with that of other methods?

  \item \textbf{RQ4:} How does AutoTraj's performance vary with different dataset sizes?

\end{itemize}




\subsection{Datasets and Experimental Setup}
\textbf{Dataset.} 
In training, as described in Section~\ref{section-4-1-1}, we sample 10,000 question–answer pairs from the Tool-Star dataset \cite{dong2025tool} and use only the questions and final answers, without any annotated trajectories. Subsequently, the Tool-Star RL dataset is adopted as the training set for the RL stage.
Besides, to comprehensively evaluate the model's tool usage capability, we conduct experiments on two types of datasets: (1) Mathematical reasoning benchmarks, including AIME2024\footnote{\url{https://huggingface.co/datasets/HuggingFaceH4/aime_2024}}, AIME2025\footnote{\url{https://huggingface.co/datasets/math-ai/aime25}}, AMC23\footnote{\url{https://huggingface.co/datasets/math-ai/amc23}}, Math \cite{hendrycks2measuring}, and GSM8K \cite{cobbe2021training}. (2) Knowledge-intensive reasoning benchmarks, including HotpotQA (HQA) \cite{yang2018hotpotqa}, 2WikiMultiHopQA (2Wiki) \cite{ho2020constructing}, MuSiQue  \cite{trivedi2022musique}, and HLE \cite{contributors2025humanity}. These benchmarks are designed to evaluate core language modeling capabilities of models, such as logical reasoning and instruction following.

\textbf{Baselines.} Our experiments consider four categories of baselines:
(1) \textit{Backbone model}: Qwen2.5-7B-Instruct \cite{qwen2025qwen25technicalreport}.
(2) \textit{SFT-only TIR methods}: Tool-Star-SFT \cite{dong2025tool}.
(3) \textit{RL-only TIR methods}: including ToRL \cite{li2025torl}, ReSearch \cite{chen2025learning}, R1-Searcher\cite{song2025r1} and AutoTIR \cite{wei2025autotir}.
(4) \textit{SFT-RL TIR Methods}: Vanilla SFT-RL TIR and Tool-Star \cite{dong2025tool}.

\textbf{Evaluation Metrics.} For mathematical reasoning tasks, we adopt an exact match based on the final answers to evaluate correctness, where a prediction is considered correct only if it exactly matches the ground-truth value. For knowledge-intensive reasoning tasks, we use the F1 score as the evaluation metric, since the same answer may be expressed in multiple valid surface forms.
During evaluation, we restrict the available tools to the Code Interpreter and the Search Engine. For fairness, all baseline methods are evaluated under the same tool configuration.

Detailed implementation details are shown in Appendix \ref{section-details}.

\subsection{Overall Performance (RQ1)}
As shown in Table~\ref{tab:accuracy}, we compare AutoTraj with a range of representative methods on two major categories of benchmarks (i.e., Mathematical Reasoning and Knowledge-Intensive Reasoning). Overall, AutoTraj achieves the best or second-best performance in terms of average accuracy and on most individual tasks, demonstrating the overall effectiveness of the proposed approach. 
More specifically, we draw the following observations.
First, from the perspective of training paradigms, the synergy between SFT and RL is crucial. SFT-only or RL-only TIR methods can yield improvements on certain tasks, but their overall performance remains clearly limited. In contrast, SFT-RL TIR methods achieve better results on most benchmarks. This indicates a strong complementarity of the two-stage training strategy: SFT equips the model with reliable tool-use behaviors and basic reasoning capabilities, while RL further optimizes reasoning efficiency and generalization at the trajectory level.
Second, compared with other SFT-RL-based reasoning methods, AutoTraj remains highly competitive across multiple tasks, particularly in terms of average accuracy and cross-task stability. This demonstrates that incorporating trajectory-level rewards in the RL stage can effectively guide the model to learn more desirable TIR trajectories, thereby validating both the trajectory reward design and the overall AutoTraj framework.
Notably, Tool-Star is cold-started with SFT on a much larger dataset (54K trajectories), whereas our SFT dataset $\mathcal{D}_{\text{sft}}$ contains only about 13K trajectories constructed from repaired reasoning paths (see Table \ref{Details of Dataset} for details.). Despite relying on fewer supervised samples, AutoTraj still outperforms Tool-Star, which further highlights the effectiveness of using repaired trajectories for SFT and the quality of the learned trajectory-level supervision.
Besides, AutoTraj achieves nearly 5 times higher reasoning efficiency than Tool-Star, indicating that it can attain higher accuracy with substantially fewer tokens. For more details, please refer to section \ref{section-5-4}.





\subsection{Ablation Experiments (RQ2)}
\begin{table}[t]
\centering
\caption{Ablation study of AutoTraj on different benchmarks.}
\label{tab:ablation}
\renewcommand\arraystretch{1.25}
\setlength{\tabcolsep}{2.2mm}{
\scalebox{0.85}{
\setlength{\aboverulesep}{0.ex}
\setlength{\belowrulesep}{0.ex}
\begin{tabular}{lccccc}
\toprule
\textbf{Method} &
\textbf{AIME24} &
\textbf{GSM8K} &
\textbf{2Wiki} &
\textbf{HLE} &
\textbf{Avg.} \\
\midrule
AutoTraj & \textbf{23.33} & \textbf{79.98} & \textbf{29.50} & \textbf{7.00} & \textbf{34.95} \\
w/o RL   & 3.33  & 77.63 & 28.00 & 6.40 & 28.84 \\
w/o RT   & 6.67  & 75.82 & 26.50 & 6.10 & 28.77 \\
w/o TR   & 3.33  & 78.85 & 28.50 & 5.20 & 28.97 \\
\bottomrule
\end{tabular}
}}
\vspace{-0.4cm}
\end{table}
To evaluate the contributions of individual modules in AutoTraj, we conduct ablation experiments summarized in Table~\ref{tab:ablation}. Specifically, we design the following comparisons. (\textit{i}) We remove the RL stage to examine the effectiveness of the high-quality SFT dataset. Under the SFT-only setting (w/o RL), the model achieves improved performance compared to Tool-Star-SFT. This demonstrates that our proposed “\textit{synthesis–filter–repair}” pipeline substantially enhances the quality of supervision, providing a stronger foundation for subsequent reasoning learning.
(\textit{ii}) We further remove the repair trajectory step in the SFT stage (w/o RT), i.e., performing SFT using only the initial high-quality trajectories, similar to previous synthetic SFT approaches. This setting is equivalent to the vanilla SFT–RL TIR approach. The results show a significant performance drop compared with AutoTraj w/o RL, indicating that repairing trajectories effectively improves model performance by correcting low-quality reasoning paths and increasing the diversity of SFT data. Specific examples of trajectory repair are provided in Appendix \ref{case study of Trajectory Repair}.
(\textit{iii})  We retain both the SFT and RL stages but remove the trajectory-level reward in the RL stage (w/o TR). The results reveal a clear performance degradation, suggesting that incorporating trajectory rewards effectively guides the model toward more accurate reasoning. 
Training curves in Appendix \ref{Reward Score} further confirm the contribution of trajectory-level rewards.

In summary, the ablation studies demonstrate that each component of AutoTraj contributes meaningfully to overall performance, with all modules playing indispensable roles within the framework.

\begin{figure}[t]
  \centering 
  \includegraphics[width = 7.8cm]{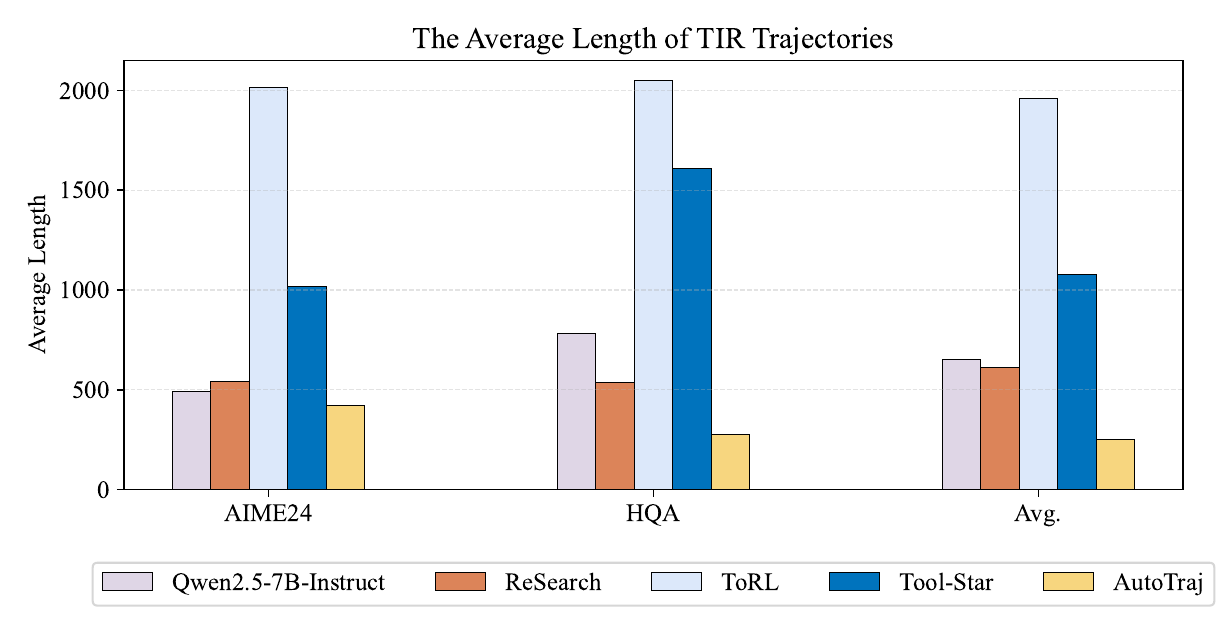}
 \caption{
Average length of TIR trajectories across benchmarks. Lower values indicate higher reasoning efficiency.
 }
 \label{avglength-1}
\vspace{-0.7cm}
\end{figure}
\subsection{Efficiency of TIR Trajectories (RQ3)}
\label{section-5-4}
In this section, we evaluate the efficiency of AutoTraj in practical reasoning tasks. Since all baseline methods are fine-tuned from the same Qwen2.5-7B model with identical parameter counts, we measure reasoning efficiency by comparing the average lengths of TIR trajectories. As shown in Figure \ref{avglength-1}, we present the average reasoning trajectory lengths of AutoTraj and several baseline methods on the AIME24 and HQA datasets, as well as the average lengths across all 9 datasets. The results indicate that AutoTraj generates significantly shorter trajectories than most baselines, demonstrating higher reasoning efficiency.
Specifically, compared with RL-only TIR methods, AutoTraj exhibits a clear advantage. For instance, ToRL has an average trajectory length of 1,960 tokens across all datasets, reflecting substantial redundancy and over-exploration during reasoning. In contrast, AutoTraj requires only 251 tokens on average, reducing trajectory length by more than a factor of seven. This dramatic reduction greatly improves reasoning efficiency while lowering computational overhead.
Among TIR methods that combine SFT and RL, AutoTraj still achieves superior trajectory efficiency. For example, although Tool-Star demonstrates competitive performance on some datasets, its average trajectory length is around 1,000 tokens, significantly longer than that of AutoTraj, resulting in lower reasoning efficiency. This indicates that AutoTraj can maintain strong reasoning performance while producing more compact and stable trajectories, further enhancing efficiency.

Overall, the results show that AutoTraj learns efficient reasoning patterns by focusing on key reasoning steps and suppressing unnecessary redundant exploration, thereby generating concise, well-structured, and effective reasoning trajectories. It achieves a favorable balance between reasoning efficiency and overall performance.
In Appendix \ref{trajectory-efficiency}, we report the average TIR trajectory lengths for all methods across all datasets, as well as the average lengths for cases where the correct answer is generated. These results further confirm the superior reasoning efficiency of AutoTraj.

\subsection{Impact of RL Training Data Scale (RQ4)}
\begin{figure}[t]
  \centering 
  
  \includegraphics[width = 7.cm]{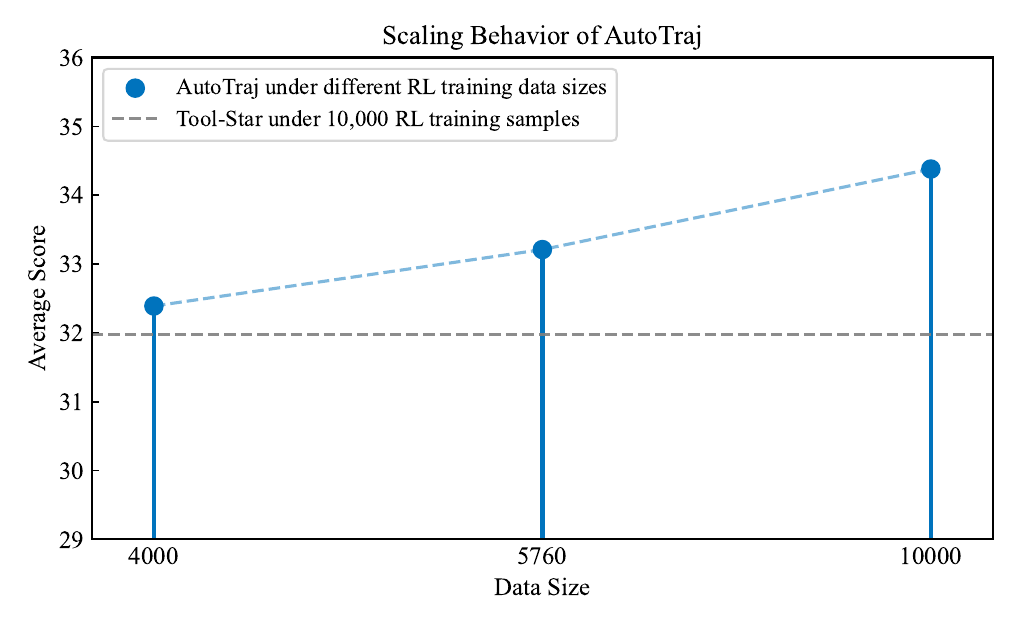}
  \vspace{-0.2cm}
 \caption{
  Average performance of AutoTraj under different RL training data scales.
 }
 \label{avgcomp-1}
\vspace{-0.4cm}
\end{figure}

In this section, we investigate the impact of RL training data scale on the performance of AutoTraj in Figure~\ref{avgcomp-1}, with the goal of assessing the effectiveness of the proposed trajectory reward in alleviating reward sparsity in RL and improving data efficiency. As shown in Figure~\ref{avgcomp-1}, during the RL stage, with only about 4,000 RL training samples, AutoTraj already achieves accuracy comparable to Tool-Star on multiple benchmarks, indicating that the trajectory reward provides denser and more informative optimization signals, enabling the model to learn high-quality TIR strategies even under limited training data. Moreover, as the training dataset size gradually increases, the overall performance of AutoTraj consistently and steadily improves. This trend suggests that the proposed trajectory reward design exhibits good scalability and can continuously guide the model to optimize its reasoning behavior as more training data become available.
In summary, these results show that AutoTraj achieves strong data efficiency in low-resource settings while maintaining stable performance gains in large-scale training regimes, highlighting its  practical applicability.


\section{Conclusion}
In this paper, we proposed AutoTraj, a two-stage framework for tool-integrated reasoning that eliminated the reliance on high-quality synthesized trajectories selected by scoring functions and mitigated reward sparsity. Specifically, in the supervised fine-tuning stage, AutoTraj automatically synthesized, evaluated, and repaired tool-integrated reasoning trajectories, providing scalable and high-quality supervision and enabling an effective cold start. In the reinforcement learning stage, AutoTraj leveraged pairs of original low-quality trajectories and their repaired counterparts to train a trajectory reward model, which supplied fine-grained reward signals during optimization. This trajectory-aware reward design explicitly accounted for reasoning quality and effectively alleviated reward sparsity. Extensive experimental results demonstrated that AutoTraj consistently improved tool-integrated reasoning performance across benchmarks, leading to more robust reasoning behaviors.

\section*{Impact Statement}
This paper presents work whose goal is to advance the field of machine learning. There are many potential societal consequences of our work, none of which we feel must be specifically highlighted here.

\bibliography{example_paper}

@article{gou2023tora,
  title={Tora: A tool-integrated reasoning agent for mathematical problem solving},
  author={Gou, Zhibin and Shao, Zhihong and Gong, Yeyun and Shen, Yelong and Yang, Yujiu and Huang, Minlie and Duan, Nan and Chen, Weizhu},
  journal={arXiv preprint arXiv:2309.17452},
  year={2023}
}

@article{li2025deepagent,
  title={DeepAgent: A General Reasoning Agent with Scalable Toolsets},
  author={Li, Xiaoxi and Jiao, Wenxiang and Jin, Jiarui and Dong, Guanting and Jin, Jiajie and Wang, Yinuo and Wang, Hao and Zhu, Yutao and Wen, Ji-Rong and Lu, Yuan and others},
  journal={arXiv preprint arXiv:2510.21618},
  year={2025}
}

@article{qian2025toolrl,
  title={Toolrl: Reward is all tool learning needs},
  author={Qian, Cheng and Acikgoz, Emre Can and He, Qi and Wang, Hongru and Chen, Xiusi and Hakkani-T{\"u}r, Dilek and Tur, Gokhan and Ji, Heng},
  journal={arXiv preprint arXiv:2504.13958},
  year={2025}
}

@article{li2023chain,
  title={Chain of code: Reasoning with a language model-augmented code emulator},
  author={Li, Chengshu and Liang, Jacky and Zeng, Andy and Chen, Xinyun and Hausman, Karol and Sadigh, Dorsa and Levine, Sergey and Fei-Fei, Li and Xia, Fei and Ichter, Brian},
  journal={arXiv preprint arXiv:2312.04474},
  year={2023}
}

@article{li2025torl,
  title={Torl: Scaling tool-integrated rl},
  author={Li, Xuefeng and Zou, Haoyang and Liu, Pengfei},
  journal={arXiv preprint arXiv:2503.23383},
  year={2025}
}

@article{chen2025learning,
  title={Learning to reason with search for llms via reinforcement learning},
  author={Chen, Mingyang and Sun, Linzhuang and Li, Tianpeng and Sun, Haoze and Zhou, Yijie and Zhu, Chenzheng and Wang, Haofen and Pan, Jeff Z and Zhang, Wen and Chen, Huajun and others},
  journal={arXiv preprint arXiv:2503.19470},
  year={2025}
}

@article{wei2025autotir,
  title={Autotir: Autonomous tools integrated reasoning via reinforcement learning},
  author={Wei, Yifan and Yu, Xiaoyan and Weng, Yixuan and Pan, Tengfei and Li, Angsheng and Du, Li},
  journal={arXiv preprint arXiv:2507.21836},
  year={2025}
}

@article{dong2025tool,
  title={Tool-Star: Empowering LLM-Brained Multi-Tool Reasoner via Reinforcement Learning},
  author={Dong, Guanting and Chen, Yifei and Li, Xiaoxi and Jin, Jiajie and Qian, Hongjin and Zhu, Yutao and Mao, Hangyu and Zhou, Guorui and Dou, Zhicheng and Wen, Ji-Rong},
  journal={arXiv preprint arXiv:2505.16410},
  year={2025}
}

@article{guo2025deepseek,
  title={Deepseek-r1: Incentivizing reasoning capability in llms via reinforcement learning},
  author={Guo, Daya and Yang, Dejian and Zhang, Haowei and Song, Junxiao and Zhang, Ruoyu and Xu, Runxin and Zhu, Qihao and Ma, Shirong and Wang, Peiyi and Bi, Xiao and others},
  journal={arXiv preprint arXiv:2501.12948},
  year={2025}
}

@article{team2024qwq,
  title={Qwq: Reflect deeply on the boundaries of the unknown},
  author={Team, Qwen},
  journal={Hugging Face},
  year={2024}
}

@article{chen2022program,
  title={Program of thoughts prompting: Disentangling computation from reasoning for numerical reasoning tasks},
  author={Chen, Wenhu and Ma, Xueguang and Wang, Xinyi and Cohen, William W},
  journal={arXiv preprint arXiv:2211.12588},
  year={2022}
}

@article{shao2024deepseekmath,
  title={Deepseekmath: Pushing the limits of mathematical reasoning in open language models},
  author={Shao, Zhihong and Wang, Peiyi and Zhu, Qihao and Xu, Runxin and Song, Junxiao and Bi, Xiao and Zhang, Haowei and Zhang, Mingchuan and Li, YK and Wu, Yang and others},
  journal={arXiv preprint arXiv:2402.03300},
  year={2024}
}

@article{wang2025steca,
  title={Steca: Step-level trajectory calibration for llm agent learning},
  author={Wang, Hanlin and Wang, Jian and Leong, Chak Tou and Li, Wenjie},
  journal={arXiv preprint arXiv:2502.14276},
  year={2025}
}

@article{wang2025answer,
  title={From< Answer> to< Think>: Multidimensional Supervision of Reasoning Process for LLM Optimization},
  author={Wang, Beining and Su, Weihang and Tian, Hongtao and Yang, Tao and Zhou, Yujia and Yao, Ting and Ai, Qingyao and Liu, Yiqun},
  journal={arXiv preprint arXiv:2510.11457},
  year={2025}
}

@article{jin2025search,
  title={Search-r1: Training llms to reason and leverage search engines with reinforcement learning},
  author={Jin, Bowen and Zeng, Hansi and Yue, Zhenrui and Yoon, Jinsung and Arik, Sercan and Wang, Dong and Zamani, Hamed and Han, Jiawei},
  journal={arXiv preprint arXiv:2503.09516},
  year={2025}
}

@article{singh2025agentic,
  title={Agentic reasoning and tool integration for llms via reinforcement learning},
  author={Singh, Joykirat and Magazine, Raghav and Pandya, Yash and Nambi, Akshay},
  journal={arXiv preprint arXiv:2505.01441},
  year={2025}
}

@article{schick2023toolformer,
  title={Toolformer: Language models can teach themselves to use tools},
  author={Schick, Timo and Dwivedi-Yu, Jane and Dess{\`\i}, Roberto and Raileanu, Roberta and Lomeli, Maria and Hambro, Eric and Zettlemoyer, Luke and Cancedda, Nicola and Scialom, Thomas},
  journal={Advances in Neural Information Processing Systems},
  volume={36},
  pages={68539--68551},
  year={2023}
}

@article{havrilla2024glore,
  title={Glore: When, where, and how to improve llm reasoning via global and local refinements},
  author={Havrilla, Alex and Raparthy, Sharath and Nalmpantis, Christoforus and Dwivedi-Yu, Jane and Zhuravinskyi, Maksym and Hambro, Eric and Raileanu, Roberta},
  journal={arXiv preprint arXiv:2402.10963},
  year={2024}
}

@article{yue2025don,
  title={Don't Overthink It: A Survey of Efficient R1-style Large Reasoning Models},
  author={Yue, Linan and Du, Yichao and Wang, Yizhi and Gao, Weibo and Yao, Fangzhou and Wang, Li and Liu, Ye and Xu, Ziyu and Liu, Qi and Di, Shimin and others},
  journal={arXiv preprint arXiv:2508.02120},
  year={2025}
}

@article{minaee2024large,
  title={Large language models: A survey},
  author={Minaee, Shervin and Mikolov, Tomas and Nikzad, Narjes and Chenaghlu, Meysam and Socher, Richard and Amatriain, Xavier and Gao, Jianfeng},
  journal={arXiv preprint arXiv:2402.06196},
  year={2024}
}

@article{paranjape2023art,
  title={Art: Automatic multi-step reasoning and tool-use for large language models},
  author={Paranjape, Bhargavi and Lundberg, Scott and Singh, Sameer and Hajishirzi, Hannaneh and Zettlemoyer, Luke and Ribeiro, Marco Tulio},
  journal={arXiv preprint arXiv:2303.09014},
  year={2023}
}

@inproceedings{yao2022react,
  title={React: Synergizing reasoning and acting in language models},
  author={Yao, Shunyu and Zhao, Jeffrey and Yu, Dian and Du, Nan and Shafran, Izhak and Narasimhan, Karthik R and Cao, Yuan},
  booktitle={The eleventh international conference on learning representations},
  year={2022}
}

@article{song2025r1,
  title={R1-searcher: Incentivizing the search capability in llms via reinforcement learning},
  author={Song, Huatong and Jiang, Jinhao and Min, Yingqian and Chen, Jie and Chen, Zhipeng and Zhao, Wayne Xin and Fang, Lei and Wen, Ji-Rong},
  journal={arXiv preprint arXiv:2503.05592},
  year={2025}
}

@article{hosain2025xolver,
  title={Xolver: Multi-Agent Reasoning with Holistic Experience Learning Just Like an Olympiad Team},
  author={Hosain, Md Tanzib and Rahman, Salman and Morol, Md Kishor and Parvez, Md Rizwan},
  journal={arXiv preprint arXiv:2506.14234},
  year={2025}
}

@inproceedings{3ZWCCELKKKF025,
  author       = {Thomas Zeng and
                  Shuibai Zhang and
                  Shutong Wu and
                  Christian Classen and
                  Daewon Chae and
                  Ethan Ewer and
                  Minjae Lee and
                  Heeju Kim and
                  Wonjun Kang and
                  Jackson Kunde and
                  Ying Fan and
                  Jungtaek Kim and
                  Hyung Il Koo and
                  Kannan Ramchandran and
                  Dimitris Papailiopoulos and
                  Kangwook Lee},
  title        = {VersaPRM: Multi-Domain Process Reward Model via Synthetic Reasoning
                  Data},
  booktitle    = {Forty-second International Conference on Machine Learning, {ICML}
                  2025, Vancouver, BC, Canada, July 13-19, 2025},
  publisher    = {OpenReview.net},
  year         = {2025},
}

@article{deng2025atom,
  title={Atom-searcher: Enhancing agentic deep research via fine-grained atomic thought reward},
  author={Deng, Yong and Wang, Guoqing and Ying, Zhenzhe and Wu, Xiaofeng and Lin, Jinzhen and Xiong, Wenwen and Dai, Yuqin and Yang, Shuo and Zhang, Zhanwei and Wang, Qiwen and others},
  journal={arXiv preprint arXiv:2508.12800},
  year={2025}
}

@inproceedings{lightman2023let,
  title={Let's verify step by step},
  author={Lightman, Hunter and Kosaraju, Vineet and Burda, Yuri and Edwards, Harrison and Baker, Bowen and Lee, Teddy and Leike, Jan and Schulman, John and Sutskever, Ilya and Cobbe, Karl},
  booktitle={The Twelfth International Conference on Learning Representations},
  year={2023}
}

@inproceedings{hendrycks2measuring,
  title={Measuring Mathematical Problem Solving With the MATH Dataset},
  author={Hendrycks, Dan and Burns, Collin and Kadavath, Saurav and Arora, Akul and Basart, Steven and Tang, Eric and Song, Dawn and Steinhardt, Jacob},
  booktitle={Thirty-fifth Conference on Neural Information Processing Systems Datasets and Benchmarks Track (Round 2)},
  year={2021}
}

@article{cobbe2021training,
  title={Training verifiers to solve math word problems},
  author={Cobbe, Karl and Kosaraju, Vineet and Bavarian, Mohammad and Chen, Mark and Jun, Heewoo and Kaiser, Lukasz and Plappert, Matthias and Tworek, Jerry and Hilton, Jacob and Nakano, Reiichiro and others},
  journal={arXiv preprint arXiv:2110.14168},
  year={2021}
}

@inproceedings{yang2018hotpotqa,
  title={HotpotQA: A dataset for diverse, explainable multi-hop question answering},
  author={Yang, Zhilin and Qi, Peng and Zhang, Saizheng and Bengio, Yoshua and Cohen, William and Salakhutdinov, Ruslan and Manning, Christopher D},
  booktitle={Proceedings of the 2018 conference on empirical methods in natural language processing},
  pages={2369--2380},
  year={2018}
}

@article{ho2020constructing,
  title={Constructing a multi-hop qa dataset for comprehensive evaluation of reasoning steps},
  author={Ho, Xanh and Nguyen, Anh-Khoa Duong and Sugawara, Saku and Aizawa, Akiko},
  journal={arXiv preprint arXiv:2011.01060},
  year={2020}
}

@article{trivedi2022musique,
  title={MuSiQue: Multihop Questions via Single-hop Question Composition},
  author={Trivedi, Harsh and Balasubramanian, Niranjan and Khot, Tushar and Sabharwal, Ashish},
  journal={Transactions of the Association for Computational Linguistics},
  volume={10},
  pages={539--554},
  year={2022},
  publisher={MIT Press One Broadway, 12th Floor, Cambridge, Massachusetts 02142, USA~…}
}

@article{contributors2025humanity,
  title={Humanity’s Last Exam},
  author={Contributors, Dataset},
  journal={arXiv preprint arXiv:2501.14249},
  year={2025}
}

@misc{qwen2025qwen25technicalreport,
      title={Qwen2.5 Technical Report}, 
      author={Qwen and : and An Yang and Baosong Yang and Beichen Zhang and Binyuan Hui and Bo Zheng and Bowen Yu and Chengyuan Li and Dayiheng Liu and Fei Huang and Haoran Wei and Huan Lin and Jian Yang and Jianhong Tu and Jianwei Zhang and Jianxin Yang and Jiaxi Yang and Jingren Zhou and Junyang Lin and Kai Dang and Keming Lu and Keqin Bao and Kexin Yang and Le Yu and Mei Li and Mingfeng Xue and Pei Zhang and Qin Zhu and Rui Men and Runji Lin and Tianhao Li and Tianyi Tang and Tingyu Xia and Xingzhang Ren and Xuancheng Ren and Yang Fan and Yang Su and Yichang Zhang and Yu Wan and Yuqiong Liu and Zeyu Cui and Zhenru Zhang and Zihan Qiu},
      year={2025},
      eprint={2412.15115},
      archivePrefix={arXiv},
      primaryClass={cs.CL},
      url={https://arxiv.org/abs/2412.15115}, 
}

@article{dong2025agentic,
  title={Agentic reinforced policy optimization},
  author={Dong, Guanting and Mao, Hangyu and Ma, Kai and Bao, Licheng and Chen, Yifei and Wang, Zhongyuan and Chen, Zhongxia and Du, Jiazhen and Wang, Huiyang and Zhang, Fuzheng and others},
  journal={arXiv preprint arXiv:2507.19849},
  year={2025}
}

@inproceedings{zheng2024llamafactory,
  title={LlamaFactory: Unified Efficient Fine-Tuning of 100+ Language Models},
  author={Yaowei Zheng and Richong Zhang and Junhao Zhang and Yanhan Ye and Zheyan Luo and Zhangchi Feng and Yongqiang Ma},
  booktitle={Proceedings of the 62nd Annual Meeting of the Association for Computational Linguistics (Volume 3: System Demonstrations)},
  address={Bangkok, Thailand},
  publisher={Association for Computational Linguistics},
  year={2024},
  url={http://arxiv.org/abs/2403.13372}
}

@article{sheng2024hybridflow,
  title   = {HybridFlow: A Flexible and Efficient RLHF Framework},
  author  = {Guangming Sheng and Chi Zhang and Zilingfeng Ye and Xibin Wu and Wang Zhang and Ru Zhang and Yanghua Peng and Haibin Lin and Chuan Wu},
  year    = {2024},
  journal = {arXiv preprint arXiv: 2409.19256}
}

@article{feng2025retool,
  title={Retool: Reinforcement learning for strategic tool use in llms},
  author={Feng, Jiazhan and Huang, Shijue and Qu, Xingwei and Zhang, Ge and Qin, Yujia and Zhong, Baoquan and Jiang, Chengquan and Chi, Jinxin and Zhong, Wanjun},
  journal={arXiv preprint arXiv:2504.11536},
  year={2025}
}

@article{kumar2025llm,
  title={Llm post-training: A deep dive into reasoning large language models},
  author={Kumar, Komal and Ashraf, Tajamul and Thawakar, Omkar and Anwer, Rao Muhammad and Cholakkal, Hisham and Shah, Mubarak and Yang, Ming-Hsuan and Torr, Phillip HS and Khan, Fahad Shahbaz and Khan, Salman},
  journal={arXiv preprint arXiv:2502.21321},
  year={2025}
}

@inproceedings{baek2025researchagent,
  title={Researchagent: Iterative research idea generation over scientific literature with large language models},
  author={Baek, Jinheon and Jauhar, Sujay Kumar and Cucerzan, Silviu and Hwang, Sung Ju},
  booktitle={Proceedings of the 2025 Conference of the Nations of the Americas Chapter of the Association for Computational Linguistics: Human Language Technologies (Volume 1: Long Papers)},
  pages={6709--6738},
  year={2025}
}

@inproceedings{acikgoz2025can,
  title={Can a single model master both multi-turn conversations and tool use? coalm: A unified conversational agentic language model},
  author={Acikgoz, Emre Can and Greer, Jeremiah and Datta, Akul and Yang, Ze and Zeng, William and Elachqar, Oussama and Koukoumidis, Emmanouil and Hakkani-Tur, Dilek and Tur, Gokhan},
  booktitle={Proceedings of the 63rd Annual Meeting of the Association for Computational Linguistics (Volume 1: Long Papers)},
  pages={12370--12390},
  year={2025}
}

@article{inoue2025drugagent,
  title={Drugagent: Multi-agent large language model-based reasoning for drug-target interaction prediction},
  author={Inoue, Yoshitaka and Song, Tianci and Wang, Xinling and Luna, Augustin and Fu, Tianfan},
  journal={ArXiv},
  pages={arXiv--2408},
  year={2025}
}

@article{hurst2024gpt,
  title={Gpt-4o system card},
  author={Hurst, Aaron and Lerer, Adam and Goucher, Adam P and Perelman, Adam and Ramesh, Aditya and Clark, Aidan and Ostrow, AJ and Welihinda, Akila and Hayes, Alan and Radford, Alec and others},
  journal={arXiv preprint arXiv:2410.21276},
  year={2024}
}

@inproceedings{FlashRAG,
  author       = {Jiajie Jin and
                  Yutao Zhu and
                  Zhicheng Dou and
                  Guanting Dong and
                  Xinyu Yang and
                  Chenghao Zhang and
                  Tong Zhao and
                  Zhao Yang and
                  Ji{-}Rong Wen},
  editor       = {Guodong Long and
                  Michale Blumestein and
                  Yi Chang and
                  Liane Lewin{-}Eytan and
                  Zi Helen Huang and
                  Elad Yom{-}Tov},
  title        = {FlashRAG: {A} Modular Toolkit for Efficient Retrieval-Augmented Generation
                  Research},
  booktitle    = {Companion Proceedings of the {ACM} on Web Conference 2025, {WWW} 2025,
                  Sydney, NSW, Australia, 28 April 2025 - 2 May 2025},
  pages        = {737--740},
  publisher    = {{ACM}},
  year         = {2025},
  url          = {https://doi.org/10.1145/3701716.3715313},
  doi          = {10.1145/3701716.3715313}
}
\bibliographystyle{icml2026}

\newpage
\appendix
\onecolumn

\section{Implementation Details}
\label{section-details}
\subsection{Datasets} 
During training, in the SFT stage, we sample 10,000 question–answer pairs from the Tool-Star-54K\footnote{\url{https://huggingface.co/datasets/dongguanting/Tool-Star-SFT-54K}} dataset and use only the questions and final answers, without any annotated trajectories, as the training data. In the RL stage, we adopt the Multi-Tool-RL-10K\footnote{\url{https://huggingface.co/datasets/dongguanting/Multi-Tool-RL-10K}} dataset as the training set for reinforcement learning.

During evaluation, we conduct experiments on two types of datasets:

(1) Mathematical reasoning benchmarks:
\begin{itemize}
  \item AIME2024\footnote{\url{https://huggingface.co/datasets/HuggingFaceH4/aime_2024}}: This dataset consists of 30 problems from the 2024 AIME I and AIME II tests. It is designed to evaluate the advanced mathematical reasoning of large language models, featuring high-difficulty problems that require multi-step solutions and unique answers. These problems test logical reasoning, mathematical understanding, and problem-solving skills. The dataset provides a challenging benchmark for assessing a model's ability to handle complex, structured reasoning tasks and generate accurate solutions.
  \item AIME2025\footnote{\url{https://huggingface.co/datasets/math-ai/aime25}}: This dataset is a collection of mathematical problems sourced from the 2025 AIME I and AIME II exams. It aims to challenge large language models with multi-step reasoning tasks that require advanced mathematical problem-solving skills. This dataset consists of a range of high-difficulty problems, carefully selected to assess a model's ability to handle complex logic, algebraic manipulation, and geometrical reasoning. It serves as a robust benchmark for evaluating the problem-solving capabilities of LLMs in the context of mathematics.
  \item AMC23\footnote{\url{https://huggingface.co/datasets/math-ai/amc23}}: This dataset is a dataset designed for evaluating mathematical reasoning and problem-solving capabilities, derived from the American Mathematics Competitions 2023. It is primarily used to assess a model’s ability to understand, reason about, and compute solutions to medium- to high-difficulty mathematical problems.
  \item Math \cite{hendrycks2measuring}: This dataset contains a collection of challenging high school-level mathematics problems, covering a wide range of topics including Prealgebra, Algebra, Number Theory, Probability, Geometry, Intermediate Algebra, and Precalculus.
  \item GSM8K \cite{cobbe2021training}: This dataset is a benchmark used to evaluate the mathematical reasoning capabilities of large language models by presenting grade-school level word problems. 
\end{itemize}

(2) Knowledge-intensive reasoning benchmarks:
\begin{itemize}
  \item HotpotQA \cite{yang2018hotpotqa}: This dataset is a multi-hop question answering dataset where the questions are based on Wikipedia articles, designed to evaluate a model's ability to integrate multiple sources of information to answer complex questions.
  \item 2WikiMultiHopQA \cite{ho2020constructing}: This dataset focuses on multi-hop reasoning, where each question requires retrieving and combining information from several distinct Wikipedia articles to provide an accurate answer.
  \item MuSiQue \cite{trivedi2022musique}: This dataset is designed for evaluating multi-step reasoning abilities, focusing on questions that require logical reasoning across multiple steps to arrive at the correct answer.
  \item HLE \cite{contributors2025humanity}: This dataset aimed at assessing the human-like reasoning and generalization capabilities of large language models. It consists of questions that challenge the models to exhibit deep reasoning, critical thinking, and the ability to provide nuanced answers.

\end{itemize}

\subsection{Details of $\mathcal{D}_{\text{sft}}$ and $\mathcal{D}_{\text{self}}$}

In this study, $\mathcal{D}_{\text{sft}}$ and $\mathcal{D}_{\text{self}}$ are two important datasets, used respectively for the SFT stage and the trajectory reward model training stage. Table~\ref{Details of Dataset} presents the statistical information of these two datasets:
\begin{itemize}
    \item The $\mathcal{D}_{\text{sft}}$ dataset contains 13,116 samples, including the Answer Correct High Quality Trajectory, Repaired Answer Correct Low Quality Trajectory, and Repaired Answer Wrong Low Quality Trajectory.
    \item The $\mathcal{D}_{\text{self}}$ dataset contains 7,178 samples, which include the Repaired Answer Correct Low Quality Trajectory, Repaired Answer Wrong Low Quality Trajectory, as well as their corresponding low-quality trajectories, i.e., Answer Correct Low Quality Trajectory, and Answer Wrong Low Quality Trajectory. These high-quality and low-quality trajectories together form the final $\mathcal{D}_{\text{self}}$ dataset.
\end{itemize}
By utilizing these datasets, AutoTraj can be effectively trained at different stages, improving both reasoning performance and trajectory quality.

\begin{table}[ht]
\centering
\caption{Dataset Statistics of the constructed SFT datasets $\mathcal{D}_{\text{sft}}$ and self-supervised datasets $\mathcal{D}_{\text{self}}$.
}
\begin{tabular}{l c}
\hline
\textbf{Dataset} & \textbf{Counts} \\
\hline
Candidate QA Pair & 10,000 \\
Candidate Tool-Integrated Trajectory & 20,000 \\
Answer Correct Trajectory & 8,413 \\
Answer Correct High Quality Trajectory & 5,938 \\
Answer Correct Low Quality Trajectory & 2,475 \\
Answer Wrong Low Quality Trajectory & 11,587 \\
Repaired Answer Correct Low Quality Trajectory & 1,491 \\
Repaired Answer Wrong Low Quality Trajectory & 5,687 \\
$\mathcal{D}_{\text{sft}}$ & 13,116 \\
$\mathcal{D}_{\text{self}}$ & 7,178 \\
\hline
\end{tabular}
\label{Details of Dataset}
\end{table}

\begin{figure}[h]
  \centering 
  \includegraphics[width = 10cm]{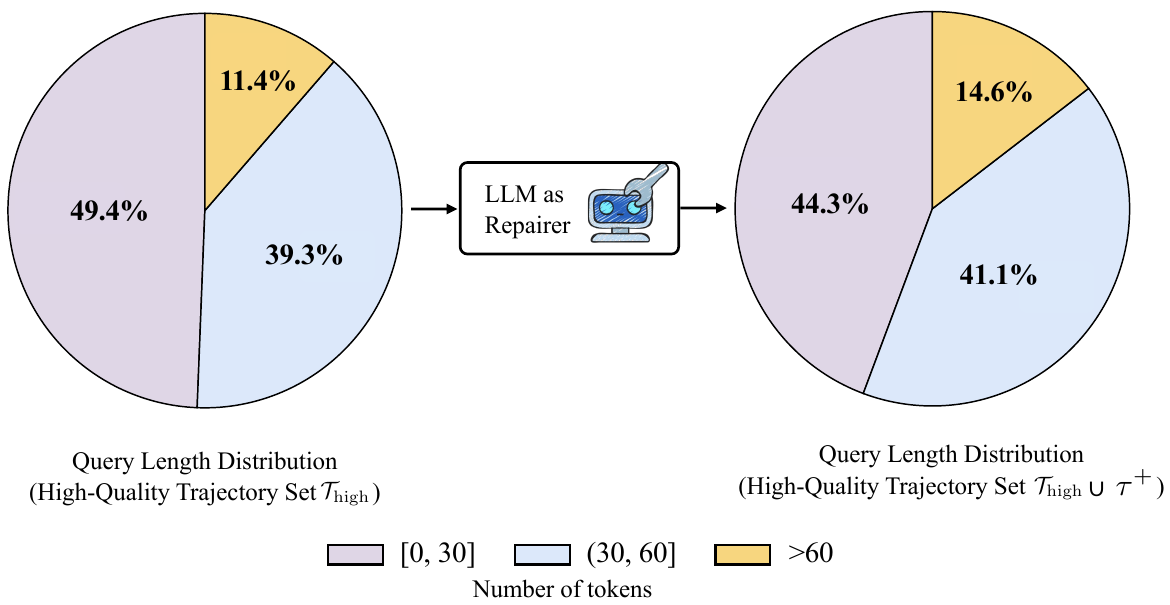}
 \caption{
  Question Length Distribution for High-Quality Trajectory Set $\mathcal{T}_\text{high}$ and High-Quality Trajectory Set $\mathcal{T}_\text{high} \cup \tau^{+}$.
 }
 \label{aclen-distribution}
\end{figure}

Besides, we conduct a analysis of the query length distribution in the High-Quality Trajectory Set $\mathcal{T}_\text{high}$ (i.e., Answer Correct High Quality Trajectory) and in High-Quality Trajectory Set $\mathcal{T}_\text{high} \cup \tau^{+}$ (i.e., $\mathcal{D}_{\text{self}}$). 
Specifically, we partition queries by length into three ranges: [0, 30], (30, 60], and $>$60 tokens. In general, longer queries are associated with higher problem difficulty.
As shown in Figure~\ref{aclen-distribution}, we observe that after introducing trajectory repair (see Section~\ref{section-4-1-3}), the proportions of medium-length ((30, 60]) and long ($>$60) queries in High-Quality Trajectory Set $\mathcal{T}_\text{high} \cup \tau^{+}$ both increase. Notably, the proportion of queries with length of $>$60 rises by as much as 3.2\%.

These results indicate that the introduction of trajectory repair leads to a noticeable shift in the query length distribution of the dataset, with a significant increase in longer questions. This shift reflects that trajectory repair effectively enhances dataset diversity, thereby laying a stronger foundation for improving model capabilities in subsequent training.

\subsection{Baselines} 

Our experiments consider four categories of baselines. In addition to the backbone model Qwen2.5-7B-Instruct, we further compare with the following three categories:

(1) SFT-only TIR methods:
\begin{itemize}
  \item Tool-Star-SFT : Tool-Star-SFT is trained on 10,000 samples (see Section~\ref{section-4-1-1}) from the Tool-Star SFT dataset based on Qwen2.5-7B-Instruct. Each sample contains not only a question–answer pair but also its corresponding reasoning trajectory.
\end{itemize}

(2) RL-only TIR methods: 
\begin{itemize}
  \item ToRL \cite{li2025torl}: ToRL is trained using reinforcement learning methods. During training, a code interpreter is integrated into the framework, allowing the model to autonomously invoke tools and more effectively tackle mathematical reasoning tasks. This framework demonstrates the potential of tool-integrated reasoning for solving mathematical problems.
  \item ReSearch \cite{chen2025learning}: ReSearch is trained using reinforcement learning methods. The framework incorporates a search mechanism for reasoning, effectively leveraging search-driven reasoning capabilities, and demonstrates its potential for real-world applications.
  \item R1-Searcher \cite{song2025r1}: R1-Searcher is a result-based two-stage reinforcement learning method designed to enhance the search capabilities of large language models. This approach enables large language models to autonomously invoke external search systems during reasoning, thereby accessing additional knowledge.
  \item AutoTIR \cite{wei2025autotir}: AutoTIR is a reinforcement learning framework that integrates a code interpreter and a search tool, enabling large language models to autonomously decide whether and which tool to invoke during the reasoning process. By utilizing a hybrid reward mechanism, AutoTIR encourages precise reasoning and effective tool integration, demonstrating versatility in its tool usage behavior.
\end{itemize}

(3) SFT-RL TIR methods: 
\begin{itemize}
\item Vanilla SFT-RL TIR: The vanilla SFT–RL TIR adopts a standard two-stage SFT–RL framework and is trained on the same AutoTraj dataset.
 Among them, in the RL stage, it optimizes the policy using only the format reward and the outcome reward.
  \item Tool-Star \cite{dong2025tool}: Tool-Star is a two-stage framework that combines SFT and reinforcement learning, designed to enable large language models to autonomously invoke multiple external tools during step-by-step reasoning. It integrates a variety of tools and incorporates thoughtful design in data synthesis and training, promoting effective tool collaboration.
\end{itemize}

\subsection{Training Settings} 
\label{training-setting}
In this paper, we use Qwen2.5-7B-Instruct as the backbone model throughout both the supervised fine-tuning (SFT) and reinforcement learning (RL) stages. 
Specifically, in Supervised Fine-Tuning via Synthesized Trajectory stage (SFT stage), our goal is to generate reasoning traces that explicitly involve tool invocation. We initially attempted to obtain such trajectories by prompt-based generation using base models such as Qwen2.5-7B-Instruct. However, we found that instruction following was often unreliable, and the generated responses frequently ignored the tool-usage protocol, producing trajectories without any tool calls. To reduce the difficulty of trajectory synthesis and ensure structurally valid tool usage, we reproduce the AutoTIR model using reinforcement learning and employ it as a dedicated trajectory generator ($f_\psi(\cdot)$), which is able to consistently produce tool-integrated reasoning trajectories. Due to computational constraints, the number of sampled candidate trajectories is set to (n=2) for each query.

In the multi-dimensional trajectory evaluation module (Section~\ref{section-4-1-2}), the weighting coefficients satisfy ($\lambda_1$+$\lambda_2$+$\lambda_3$=1), and we set $\lambda_1$=0.4 and $\lambda_2$=$\lambda_3$=0.3 in all experiments. 
During trajectory classification, we set two thresholds $\theta_\text{qual}$ for correct-answer trajectories, both determined through observations on small-batch trajectory scores. The first threshold, \textit{high quality threshold} $\theta^{1}_\text{qual}$, is set to 0.86. For trajectories scoring above this threshold on the same query, the highest-scoring trajectory is labeled as high-quality, while the rest are labeled as low-quality. The second threshold is \textit{all high quality threshold} $\theta^{2}_\text{qual}$, set to 0.9. If all trajectories on a given query score above this threshold, all trajectories on that query are classified as high-quality. This strategy prevents unnecessary downgrading of trajectories with high overall quality and minimal variation, thereby avoiding noise introduction during subsequent reward model training.
Ultimately, all initial trajectories are categorized into three groups: high-quality trajectories with correct answers $\mathcal{T}_\text{high}$, low-quality trajectories with correct answers $\mathcal{T}_\text{low}^{\text{correct}}$, and low-quality trajectories with incorrect answers $\mathcal{T}_\text{low}^{\text{wrong}}$. This provides a structured data foundation for subsequent supervised fine-tuning and trajectory reward model training.

For trajectory repair, we adopt Qwen2.5-32B-Instruct as the LLM-as-Repairer. From Table~\ref{Details of Dataset}, it can be observed that our repair success rate is over 50\%. For trajectories that remain low quality after the first repair, we have not yet conducted further processing in the current work to improve trajectory utilization. In the future, more advanced models could be introduced to attempt a second round of repairs, or these trajectories could be utilized in other potential use cases.

In the Trajectory-Aware Reinforcement Learning stage (RL stage), the trajectory reward model is trained based on Qwen2.5-7B-Instruct. GRPO is used for policy optimization, where the trajectory reward weight is set to $\alpha$=0.2, and the maximum reward is capped at $R_{\max}$=3. To prevent environment feedback from biasing the policy update, the execution results of tool invocations are explicitly masked during loss computation and excluded from gradient backpropagation.

The SFT stage and the training of the trajectory reward model are implemented using the LLaMA Factory framework~\cite{zheng2024llamafactory}, while the GRPO-based reinforcement learning stage is conducted with the Verl framework~\cite{sheng2024hybridflow}. All experiments are carried out on 4 NVIDIA A800 GPUs, and detailed hyperparameter settings are provided in our released code \footnote{\url{https://github.com/GSYY0607/AutoTraj_code}}.

\subsection{Evaluation Settings} 
During evaluation, we only enable the Code Interpreter and Search tools, and all baseline methods are tested under the same tool configuration for a fair comparison:
\begin{itemize}
  \item Code: Code generated by the large language model is executed in a sandbox environment, and the corresponding execution results are returned. In implementing the Python interpreter, we followed the design approach of Tool-Star \cite{dong2025tool}.
  \item Search: The search query is executed to retrieve relevant information, utilizing the Wikipedia search service implemented based on FlashRAG ~\cite{FlashRAG}.
\end{itemize}

\section{More Experiments}

\subsection{Efficiency of TIR Trajectories}
\label{trajectory-efficiency}
In this section, Table~\ref{tab:lenall} first presents the average reasoning trajectory lengths of AutoTraj and baseline methods across all datasets. We then report in Table~\ref{tab:lencorrect} the average trajectory length when different methods successfully generate \textit{correct} answers.
It can be observed that, once correct predictions are achieved, the average trajectory length decreases for all methods. Nevertheless, compared with most baselines, AutoTraj consistently produces substantially shorter correct reasoning trajectories, indicating a clear advantage in reasoning efficiency.
Overall, these results demonstrate that AutoTraj achieves a favorable balance between reasoning efficiency and overall performance.

\begin{table*}[t]
\centering
\caption{Average reasoning trajectory lengths of AutoTraj and baseline methods across all benchmarks, where shorter trajectories indicate better performance. The top two results are highlighted in \textbf{bold} and \underline{underlined}.}
  \renewcommand\arraystretch{1.25}
\setlength{\tabcolsep}{2.4mm}{
    \scalebox{0.81}{
    \setlength{\aboverulesep}{0.ex}
\setlength{\belowrulesep}{0.ex}
\begin{tabular}{lccccccccccc}
\toprule
\hline
\multirow{2.2}{*}{\textbf{Method}} 
& \multicolumn{5}{c}{\textbf{Mathematical Reasoning}} 
& \multicolumn{4}{c}{\textbf{Knowledge-Intensive Reasoning}} 
& \multirow{2.2}{*}{\textbf{Avg.}} \\
\cmidrule(lr){2-6} \cmidrule(lr){7-10} \addlinespace[-1.1em]\\
 & AIME24 & AIME25 & AMC23 & GSM8K & MATH & HQA & 2Wiki & MuSiQue & HLE &  \\
\midrule
Qwen2.5-7B-Instruct             & 491.53 & 1,094.33 & 414.62 & 519.61 & 410.46 & 780.11 & 1,033.64 & 1,117.68 & 876.89 & 652.69 \\
\midrule
\multicolumn{10}{l}{\textit{\textbf{SFT-only TIR Methods}}} \\
Tool-Star-SFT                   & 1,172.77 & 1,269.10 & 830.85 & 255.88 & 337.02 & 1,273.44 & 1,003.94 & 1,513.54 & 1,459.48 & 746.13\\
\midrule
\multicolumn{10}{l}{\textit{\textbf{RL-only TIR Methods}}} \\
ReSearch                        & 540.37 & 1,001.80 & 611.00 & 498.62 & 463.56 & 535.73 & 445.89 & 676.87 & 933.09 & 611.28 \\
R1-Searcher                     & 1,027.93 & 949.67 & 908.67 & 554.45 & 510.69 & 597.88 & 405.63 & 533.92 & 780.82 & 603.63 \\
ToRL                            & 2,014.57 & 1,885.57 & 1,836.75 & 2,071.49 & 1,794.49 & 2,048.26 & 1,873.03 & 2,178.26 & 1,939.53 & 1,959.87 \\
AutoTIR                         & \textbf{315.03} & \textbf{301.80} & \textbf{235.88} & \textbf{156.30} & \textbf{160.30} & \underline{342.68} & \underline{349.83} & \underline{446.56} & \textbf{326.82} & \textbf{236.14} \\
\midrule
\multicolumn{10}{l}{\textit{\textbf{SFT-RL TIR Methods}}} \\
Vanilla SFT–RL TIR              & 789.13 & 746.33 & 655.83 & 496.45 & 498.08 & 404.30 & 405.67 & 455.69 & 656.96 & 531.30 \\
Tool-Star                       & 1,019.40 & 1,059.80 & 1,120.05 & 573.82 & 646.91 & 1,607.88 & 1,308.56 & 1,973.41 & 1,843.90 & 1,078.90 \\
\rowcolor{bestcell!90} AutoTraj & \underline{423.17} & \underline{433.97} & \underline{306.40} & \underline{158.44} & \underline{176.10} & \textbf{276.36} & \textbf{230.03} & \textbf{382.25} & \underline{409.43} & \underline{251.36} \\
\bottomrule
\hline
\end{tabular}
    }
    }  

\label{tab:lenall}
\end{table*}

\begin{table*}[t]
\centering
\caption{Average reasoning trajectory lengths of AutoTraj and baseline methods across all benchmarks when different methods successfully generate \textit{correct} answers. Shorter trajectories indicate better performance. The top two results are highlighted in \textbf{bold} and \underline{underlined}.}
  \renewcommand\arraystretch{1.25}
\setlength{\tabcolsep}{2.4mm}{
    \scalebox{0.81}{
    \setlength{\aboverulesep}{0.ex}
\setlength{\belowrulesep}{0.ex}
\begin{tabular}{lccccccccccc}
\toprule
\hline
\multirow{2.2}{*}{\textbf{Method}} 
& \multicolumn{5}{c}{\textbf{Mathematical Reasoning}} 
& \multicolumn{4}{c}{\textbf{Knowledge-Intensive Reasoning}} 
& \multirow{2.2}{*}{\textbf{Avg.}} \\
\cmidrule(lr){2-6} \cmidrule(lr){7-10} \addlinespace[-1.1em]\\
 & AIME24 & AIME25 & AMC23 & GSM8K & MATH & HQA & 2Wiki & MuSiQue & HLE &  \\
\midrule
Qwen2.5-7B-Instruct             & 372.50 & 0.00 & 461.73 & 594.82 & 411.09 & 536.45 & 808.49 & 747.38 & 832.57 & 532.95 \\
\midrule
\multicolumn{10}{l}{\textit{\textbf{SFT-only TIR Methods}}} \\
Tool-Star-SFT                   & 462.50 & 956.25 & 594.79 & 239.24 & 275.37 & 1,299.89 & 1,140.81 & 1,604.75 & 967.02 & 355.54\\
\midrule
\multicolumn{10}{l}{\textit{\textbf{RL-only TIR Methods}}} \\
ReSearch                        & 0.00 & 1,266.00 & 417.33 & 466.76 & 414.47 & 468.57 & 395.47 & 562.03 & 876.22 & 467.39 \\
R1-Searcher                     & \textbf{180.00} & 0.00 & 627.60 & 376.11 & 372.94 & 541.10 & \underline{271.38} & \underline{306.94} & 563.05 & 381.92 \\
ToRL                            & 1,856.88 & 1,467.00 & 1,965.71 & 2,101.65 & 1,785.99 & 2,004.54 & 1,895.57 & 2,279.50 & 1,952.55 & 1,968.40 \\
AutoTIR                         & \underline{226.50} & \textbf{159.00} & \textbf{244.07} & \underline{155.14} & \textbf{143.29} & \underline{350.51} & 329.66 & 359.45 & \textbf{318.04} & \underline{175.84} \\
\midrule
\multicolumn{10}{l}{\textit{\textbf{SFT-RL TIR Methods}}} \\
Vanilla SFT–RL TIR              & 448.00 & 851.00 & 605.05 & 487.87 & 487.02 & 396.42 & 422.63 & 415.71 & 601.73 & 486.25 \\
Tool-Star                       & 735.67 & 1,267.50 & 931.10 & 457.40 & 522.89 & 1,166.54 & 1,317.69 & 1,505.12 & 2,007.91 & 576.54 \\
\rowcolor{bestcell!90} AutoTraj & 250.57 & \underline{199.67} & \underline{335.21} & \textbf{144.16} & \underline{151.81} & \textbf{260.76} & \textbf{250.04} & \textbf{292.83} & \underline{358.67} & \textbf{165.30} \\
\bottomrule
\hline
\end{tabular}
    }
    }  

\label{tab:lencorrect}
\end{table*}



\subsection{Trajectory Quality Analysis}

While answer accuracy indicates whether a model reaches the correct conclusion, it does not fully reflect the quality of the underlying reasoning process. We therefore further evaluate the quality of \textit{correct reasoning trajectories} from a multi-dimensional perspective (i.e., using Eq.(\ref{multi})). Specifically, Table~\ref{tab:avgscore} reports the results under the \emph{Multi-Dimensional Trajectory Evaluation} framework (see Section~\ref{section-4-1-2}).
Since only a single reasoning trajectory is generated for each query in this setting, all methods achieve identical scores on the \emph{length} dimension, and this factor does not contribute to performance differences. As a result, the reported scores mainly reflect trajectory quality along the \emph{confidence} and \emph{repetition} dimensions.
Across both mathematical reasoning and knowledge-intensive reasoning tasks, AutoTraj consistently obtains higher scores than most baseline methods, indicating clear advantages in the quality of its correct reasoning trajectories. In particular, AutoTraj achieves an average score of \textbf{0.88} over all datasets, substantially outperforming RL-only methods such as ToRL (0.74), and also surpassing hybrid SFT-RL approaches including Tool-Star (0.79).

This notable performance gap suggests that AutoTraj is able to complete the reasoning process in a more concise and structurally coherent manner. 
Beyond improving answer correctness, AutoTraj effectively enhances the overall quality of reasoning trajectories by reducing redundancy and improving logical clarity.

\begin{table*}[t]
\centering
\caption{Average score of AutoTraj and baseline methods across all benchmarks under the \emph{Multi-Dimensional Trajectory Evaluation} framework. The top two results are highlighted in \textbf{bold} and \underline{underlined}.}
  \renewcommand\arraystretch{1.25}
\setlength{\tabcolsep}{3.0mm}{
    \scalebox{0.81}{
    \setlength{\aboverulesep}{0.ex}
\setlength{\belowrulesep}{0.ex}
\begin{tabular}{lccccccccccc}
\toprule
\hline
\multirow{2.2}{*}{\textbf{Method}} 
& \multicolumn{5}{c}{\textbf{Mathematical Reasoning}} 
& \multicolumn{4}{c}{\textbf{Knowledge-Intensive Reasoning}} 
& \multirow{2.2}{*}{\textbf{Avg.}} \\
\cmidrule(lr){2-6} \cmidrule(lr){7-10} \addlinespace[-1.1em]\\
 & AIME24 & AIME25 & AMC23 & GSM8K & MATH & HQA & 2Wiki & MuSiQue & HLE &  \\
\midrule
Qwen2.5-7B-Instruct             & \underline{0.93} & 0.00 & \textbf{0.91} & 0.86 & \underline{0.90} & \underline{0.85} & 0.78 & 0.77 & 0.80 & 0.75 \\
\midrule
\multicolumn{10}{l}{\textit{\textbf{SFT-only TIR Methods}}} \\
Tool-Star-SFT                   & 0.80 & \underline{0.82} & 0.88 & \underline{0.92} & \textbf{0.92} & 0.77 & 0.79 & 0.73 & 0.72 & 0.82 \\
\midrule
\multicolumn{10}{l}{\textit{\textbf{RL-only TIR Methods}}} \\
ReSearch                        & 0.00 & 0.37 & 0.76 & 0.88 & \underline{0.90} & 0.82 & 0.81 & 0.76 & \underline{0.81} & 0.68 \\
R1-Searcher                     & \textbf{0.95} & 0.00 & \underline{0.89} & 0.86 & 0.88 & 0.83 & \underline{0.87} & \textbf{0.87} & \textbf{0.85} & 0.78 \\
ToRL                            & 0.74 & 0.79 & 0.75 & 0.72 & 0.73 & 0.72 & 0.73 & 0.72 & 0.74 & 0.74 \\
AutoTIR                         & 0.80 & 0.67 & 0.87 & 0.89 & \underline{0.90} & \underline{0.85} & 0.86 & 0.85 & \textbf{0.85} & \underline{0.83} \\
\midrule
\multicolumn{10}{l}{\textit{\textbf{SFT-RL TIR Methods}}} \\
Vanilla SFT–RL TIR              & 0.78 & 0.78 & 0.80 & 0.78 & 0.81 & 0.75 & 0.75 & 0.77 & 0.76 & 0.78 \\
Tool-Star                       & 0.87 & \textbf{0.84} & 0.82 & 0.85 & 0.84 & 0.74 & 0.77 & 0.63 & 0.74 & 0.79 \\
\rowcolor{bestcell!90} AutoTraj & \underline{0.93} & 0.76 & \underline{0.89} & \textbf{0.94} & \textbf{0.92} & \textbf{0.87} & \textbf{0.89} & \underline{0.86} & \textbf{0.85} & \textbf{0.88} \\
\bottomrule
\hline
\end{tabular}
    }
    }  

\label{tab:avgscore}
\end{table*}


\subsection{Reward Score Analysis}
\label{Reward Score}
\begin{figure}[t]
  \centering 
  \includegraphics[width = 13.5cm]{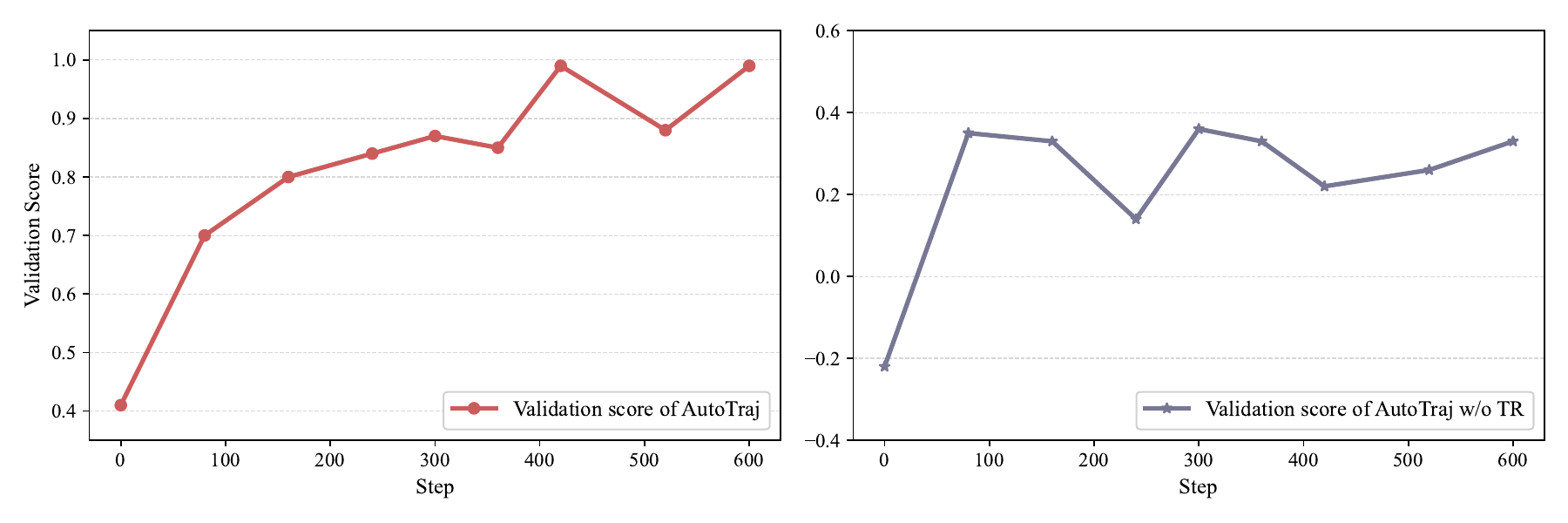}
 \caption{
Validation score of AutoTraj and its ablated variant without trajectory rewards (i.e., AutoTraj w/o TR) during GRPO training.}
 \label{valscore-1}

\end{figure}

To analyze the role of trajectory rewards in the RL stage, we examine the evolution of validation rewards during GRPO training. As shown in the left of Figure~\ref{valscore-1}, after incorporating trajectory rewards, the reward of AutoTraj exhibits a stable and steadily increasing trend over training iterations, indicating a smooth and effective improvement in reasoning capability. In contrast, the right of Figure~\ref{valscore-1} presents the training curve of AutoTraj w/o TR, where the trajectory reward is removed. We observe significant fluctuations with large variance in the reward values, a noticeably slower convergence rate, and overall unstable training dynamics. This suggests that, without trajectory-level reward signals, the optimization feedback becomes sparser and noisier, thereby weakening the learning of high-quality reasoning behavior.
Finally, this comparison further demonstrates that trajectory rewards provide denser and more stable optimization signals for reinforcement learning, which in turn substantially improve the training stability and reasoning performance of AutoTraj.

\section{System Prompt Template}
\label{prompt}
\subsection{System Prompt Template for Candidate Tool-Integrated Trajectory Synthesis}
\begin{figure*}[t]
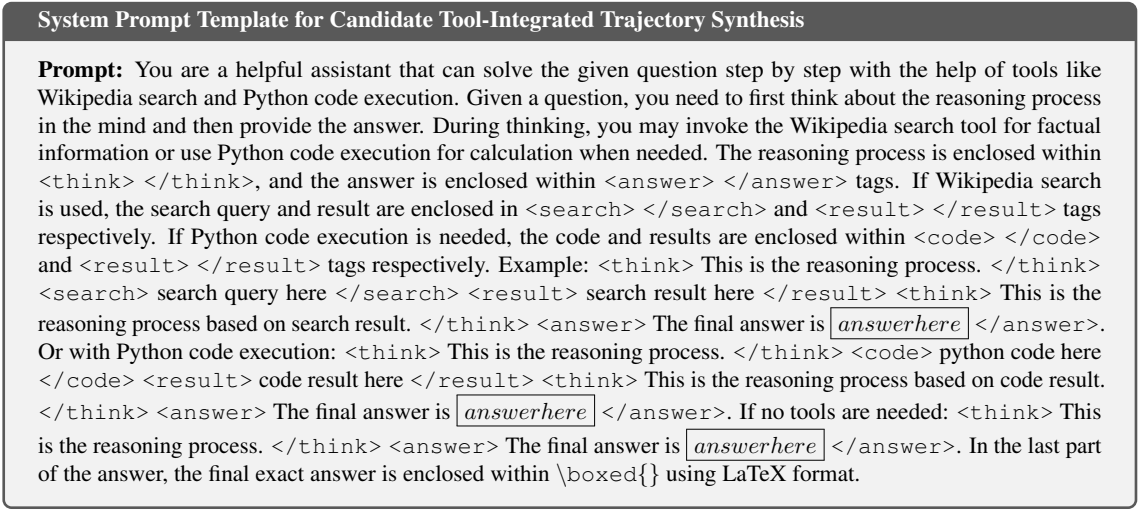

  \centering
  \scalebox{0.88}{
\begin{tcolorbox}[colframe=black!65!white, colback=black!50!white!10, coltitle=white, boxrule=0.4mm, arc=1mm, title=\textbf{System Prompt Template for Candidate Tool-Integrated Trajectory Synthesis}]
  \textbf{Prompt:} \textnormal{You are a helpful assistant that can solve the given question step by step with the help of tools like Wikipedia search and Python code execution. Given a question, you need to first think about the reasoning process in the mind and then provide the answer. During thinking, you may invoke the Wikipedia search tool for factual information or use Python code execution for calculation when needed. The reasoning process is enclosed within \texttt{<think>} \texttt{</think>}, and the answer is enclosed within \texttt{<answer>} \texttt{</answer>} tags. If Wikipedia search is used, the search query and result are enclosed in \texttt{<search>} \texttt{</search>} and \texttt{<result>} \texttt{</result>} tags respectively. If Python code execution is needed, the code and results are enclosed within \texttt{<code>} \texttt{</code>} and \texttt{<result>} \texttt{</result>} tags respectively. Example: \texttt{<think>} This is the reasoning process. \texttt{</think>} \texttt{<search>} search query here \texttt{</search>} \texttt{<result>} search result here \texttt{</result>} \texttt{<think>} This is the reasoning process based on search result. \texttt{</think>} \texttt{<answer>} The final answer is \boxed{answer here} \texttt{</answer>}. Or with Python code execution: \texttt{<think>} This is the reasoning process. \texttt{</think>} \texttt{<code>} python code here \texttt{</code>} \texttt{<result>} code result here \texttt{</result>} \texttt{<think>} This is the reasoning process based on code result. \texttt{</think>} \texttt{<answer>} The final answer is \boxed{answer here} \texttt{</answer>}. If no tools are needed: \texttt{<think>} This is the reasoning process. \texttt{</think>} \texttt{<answer>} The final answer is \boxed{answer here} \texttt{</answer>}. In the last part of the answer, the final exact answer is enclosed within \texttt{\textbackslash boxed\{\}} using LaTeX format.}
\end{tcolorbox}}
\caption{System Prompt Template for Candidate Tool-Integrated Trajectory Synthesis.}
\label{prompt-1}
\end{figure*}

 As shown in Figure~\ref{prompt-1}, the model is allowed to autonomously decide whether to invoke external tools during the reasoning process to facilitate problem solving.
\label{prompt1}

\subsection{System Prompt Template for Trajectory Repair}
\label{prompt2}
As shown in Figure~\ref{prompt-2}, for trajectories with correct answers but low quality, the prompt first guides the model to analyze the original trajectory, focusing on identifying potential issues such as redundant reasoning, logical inconsistencies, or imprecise reasoning caused by ambiguous expressions like “seems” or “recheck.” This analysis process serves solely as an internal reasoning step for the model and is not included in the final generated trajectory. Subsequently, the model regenerates a repaired trajectory based on the question while adhering to the same constraints as the initial generation phase. 
\begin{figure*}[t]
  \centering
  \scalebox{0.88}{
\begin{tcolorbox}[colframe=black!65!white, colback=black!50!white!10, coltitle=white, boxrule=0.4mm, arc=1mm, title=\textbf{System Prompt Template for Answer Correct Trajectory Repair}]
  \textbf{Prompt:} \textnormal{
You are a helpful assistant that solves questions step by step using tools. \\
You are given: \\
- A question \\
- A low-quality reasoning trajectory that reaches the correct answer \\
Before generating, silently analyze whether the old trajectory contains: \\
- redundant reasoning \\
- unclear logic \\
- vague or uncertain words (e.g., seems, maybe, perhaps, probably) \\
Do NOT write this analysis in your output. \\
Then regenerate a high-quality reasoning trajectory from scratch. \\
REQUIREMENTS: \\
1. The FIRST non-whitespace token MUST be exactly \texttt{<think>} \\
2. ALL reasoning MUST appear ONLY inside \texttt{<think>...</think>} \\
3. You MUST NOT reveal reasoning outside \texttt{<think>} tags \\
4. You MAY use tools ONLY if necessary, and AT MOST 3 tool calls \\
5. Use tools ONLY when needed. If Wikipedia search is used, the search query and result are enclosed within \texttt{<search> </search>} and \texttt{<result> </result>} tags respectively. If Python code execution is needed, the code and results are enclosed within \texttt{<code> </code>} and \texttt{<result> </result>} tags respectively. \\ 
6. You MUST reason again inside \texttt{<think>} after each \texttt{<result>} \\
7. You MUST produce EXACTLY ONE \texttt{<answer>...</answer>} block \\
8. The \texttt{<answer>} block MUST be the FINAL output. After \texttt{</answer>}, output NOTHING \\
9. You MUST NOT output multiple \texttt{<answer>} blocks \\
10. Do NOT use uncertain words like seems, maybe, perhaps, probably, might, likely \\
11. Keep reasoning concise, logical, and clear \\
12. Inside \texttt{<answer>}, output ONLY the final answer formatted exactly as: \texttt{\textbackslash boxed\{final\_answer\}}}
\end{tcolorbox}}
\caption{System Prompt Template for Answer Correct Trajectory Repair.}
\label{prompt-2}
\end{figure*}

As shown in Figure~\ref{prompt-3}, for low-quality trajectories with incorrect answers, the prompt design focuses on guiding the model to progressively analyze the root cause of the final answer error. This analysis is also not written into the generated trajectory. Based on this, the model regenerates a repair trajectory satisfying the same constraints as the initial phase. 
\begin{figure*}[t]
  \centering
  \scalebox{0.88}{
\begin{tcolorbox}[colframe=black!65!white, colback=black!50!white!10, coltitle=white, boxrule=0.4mm, arc=1mm, title=\textbf{System Prompt Template for Answer False Trajectory Repair}]
  \textbf{Prompt:} \textnormal{
You are a helpful assistant that solves questions step by step using tools. \\
You are given: \\
- A question \\
- A low-quality reasoning trajectory that leads to an INCORRECT answer \\
Before generating, silently analyze step by step: \\
- where the reasoning goes wrong \\
- whether the error is caused by misunderstanding, logic flaw, calculation mistake, or tool misuse \\
- how the mistake propagates to the final incorrect answer \\
Do NOT write this analysis in your output. \\
Then regenerate a CORRECT and high-quality reasoning trajectory from scratch. \\
REQUIREMENTS: \\
1. The FIRST non-whitespace token MUST be exactly \texttt{<think>} \\
2. ALL reasoning MUST appear ONLY inside \texttt{<think>...</think>} \\
3. You MUST NOT reveal reasoning outside \texttt{<think>} tags \\
4. You MAY use tools ONLY if necessary, and AT MOST 3 tool calls \\
5. Use tools ONLY when needed. If Wikipedia search is used, the search query and result are enclosed within \texttt{<search> </search>} and \texttt{<result> </result>} tags respectively. If Python code execution is needed, the code and results are enclosed within \texttt{<code> </code>} and \texttt{<result> </result>} tags respectively. \\
6. You MUST reason again inside \texttt{<think>} after each \texttt{<result>} \\
7. You MUST produce EXACTLY ONE \texttt{<answer>...</answer>} block \\
8. The \texttt{<answer>} block MUST be the FINAL output. After \texttt{</answer>}, output NOTHING \\
9. You MUST NOT output multiple \texttt{<answer>} blocks \\
10. Do NOT use uncertain words like seems, maybe, perhaps, probably, might, likely \\
11. Keep reasoning concise, logical, and correct \\
12. Inside \texttt{<answer>}, output ONLY the final answer formatted exactly as: \texttt{\textbackslash boxed\{final\_answer\}}}
\end{tcolorbox}}
\caption{System Prompt Template for Answer False Trajectory Repair.}
\label{prompt-3}
\end{figure*}

\subsection{System Prompt Template for GRPO}
\label{prompt3}
 The system prompt template for GRPO is shown in Figure~\ref{prompt-4}.
\begin{figure*}[t]
  \centering
  \scalebox{0.88}{
\begin{tcolorbox}[colframe=black!65!white, colback=black!50!white!10, coltitle=white, boxrule=0.4mm, arc=1mm, title=\textbf{System Prompt Template for GRPO}]
  \textbf{Prompt:} \textnormal{
You are a helpful assistant that can solve the given question step by step with the help of tools like Wikipedia search and Python code execution. Given a question, you need to first think about the reasoning process in the mind and then provide the answer. During thinking, you may invoke the Wikipedia search tool for factual information or use Python code execution for calculation when needed. The reasoning process is enclosed within \texttt{<think> </think>}, and the answer is enclosed within \texttt{<answer> </answer>} tags. If Wikipedia search is used, the search query and result are enclosed in \texttt{<search> </search>} and \texttt{<result> </result>} tags respectively. If Python code execution is needed, the code and results are enclosed within \texttt{<code> </code>} and \texttt{<result> </result>} tags respectively. Example: \texttt{<think>} This is the reasoning process. \texttt{</think>} \texttt{<search>} search query here \texttt{</search>} \texttt{<result>} search result here \texttt{</result>} \texttt{<think>} This is the reasoning process based on search result. \texttt{</think>} \texttt{<answer>} The final answer is \texttt{\textbackslash boxed\{answer here\}} \texttt{</answer>}. Or with Python code execution: \texttt{<think>} This is the reasoning process. \texttt{</think>} \texttt{<code>} python code here \texttt{</code>} \texttt{<result>} code result here \texttt{</result>} \texttt{<think>} This is the reasoning process based on code result. \texttt{</think>} \texttt{<answer>} The final answer is \texttt{\textbackslash boxed\{answer here\}} \texttt{</answer>}. If no tools are needed: \texttt{<think>} This is the reasoning process. \texttt{</think>} \texttt{<answer>} The final answer is \texttt{\textbackslash boxed\{answer here\}} \texttt{</answer>}. In the last part of the answer, the final exact answer is enclosed within \texttt{\textbackslash boxed\{\}} with LaTeX format.
}
\end{tcolorbox}}
\caption{System Prompt Template for GRPO.}
\label{prompt-4}
\end{figure*}

\section{Case Study}
\subsection{Cases of Trajectory Repair}
\label{case study of Trajectory Repair}
\begin{figure*}[t]
  \centering 
  \includegraphics[width = 15cm]{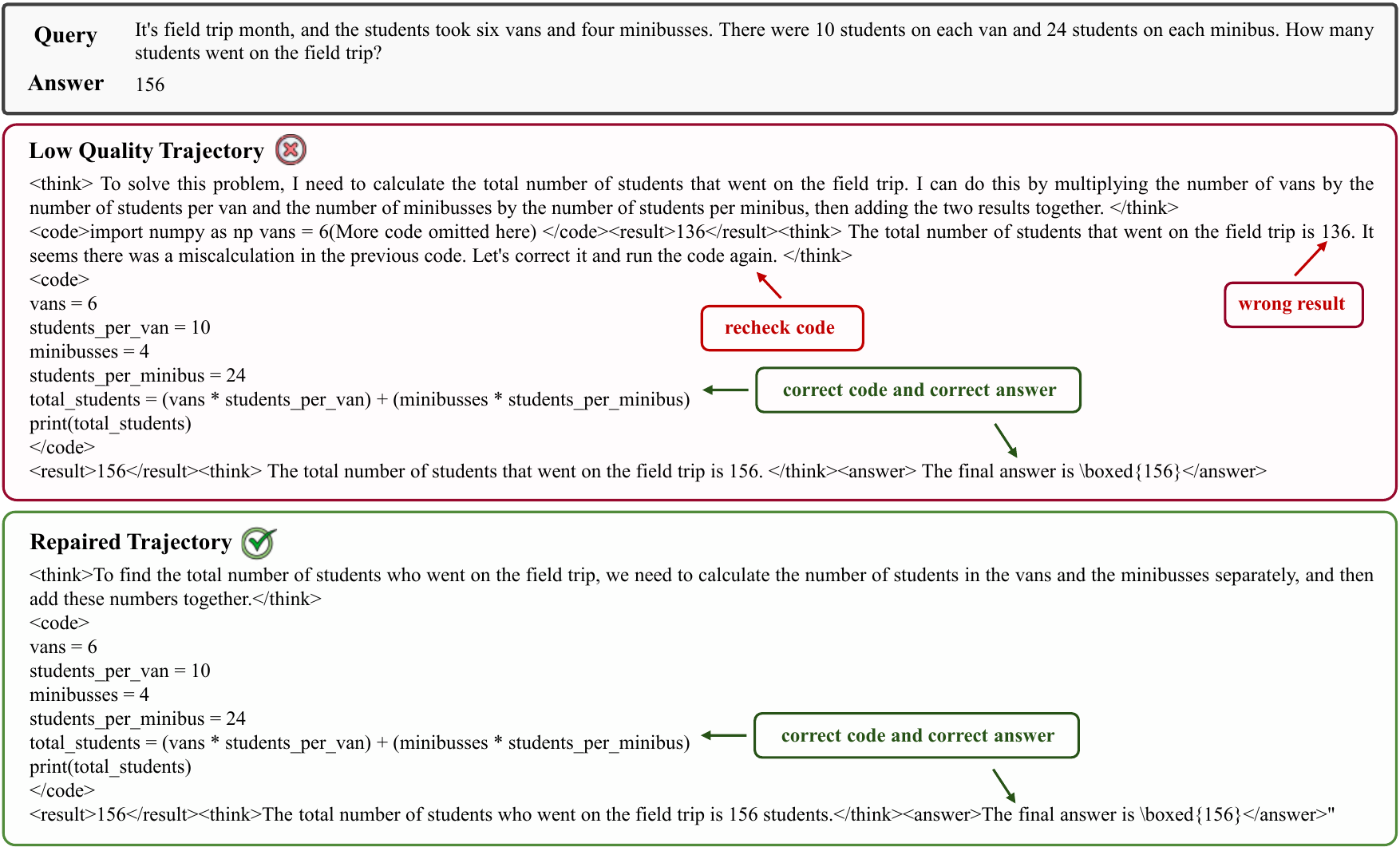}
 \caption{
  A Case study for an answer correct low quality trajectory repair. 
 }
 \label{repaircasestudy-1}

\end{figure*}

We further conduct case studies to illustrate how trajectory repair operates in practical applications. Specifically, we consider two types of repairs: Answer-Correct Low-Quality Trajectories (i.e., $\mathcal{T}_\text{low}^{\text{correct}}$) and Answer-Wrong Low-Quality Trajectories (i.e., $\mathcal{T}_\text{low}^{\text{wrong}}$). In Figure \ref{repaircasestudy-1}, we first illustrate the repair of an answer-correct low-quality trajectory. Specifically, the repairer identifies that the low-quality trajectory contains an error in which incorrect code is generated, leading to an erroneous intermediate result. Although the original model subsequently recognizes that a mistake may have occurred and regenerates the code to obtain the correct answer, this initial error still undermines the overall reliability of the reasoning process.
In the repaired trajectory, this issue is effectively eliminated. In details, the repairer generates correct code when invoking the code tool and returns the correct result, thereby reaching the correct answer in a smooth and coherent manner.

\begin{figure*}[t]
  \centering 
  \includegraphics[width = 15cm]{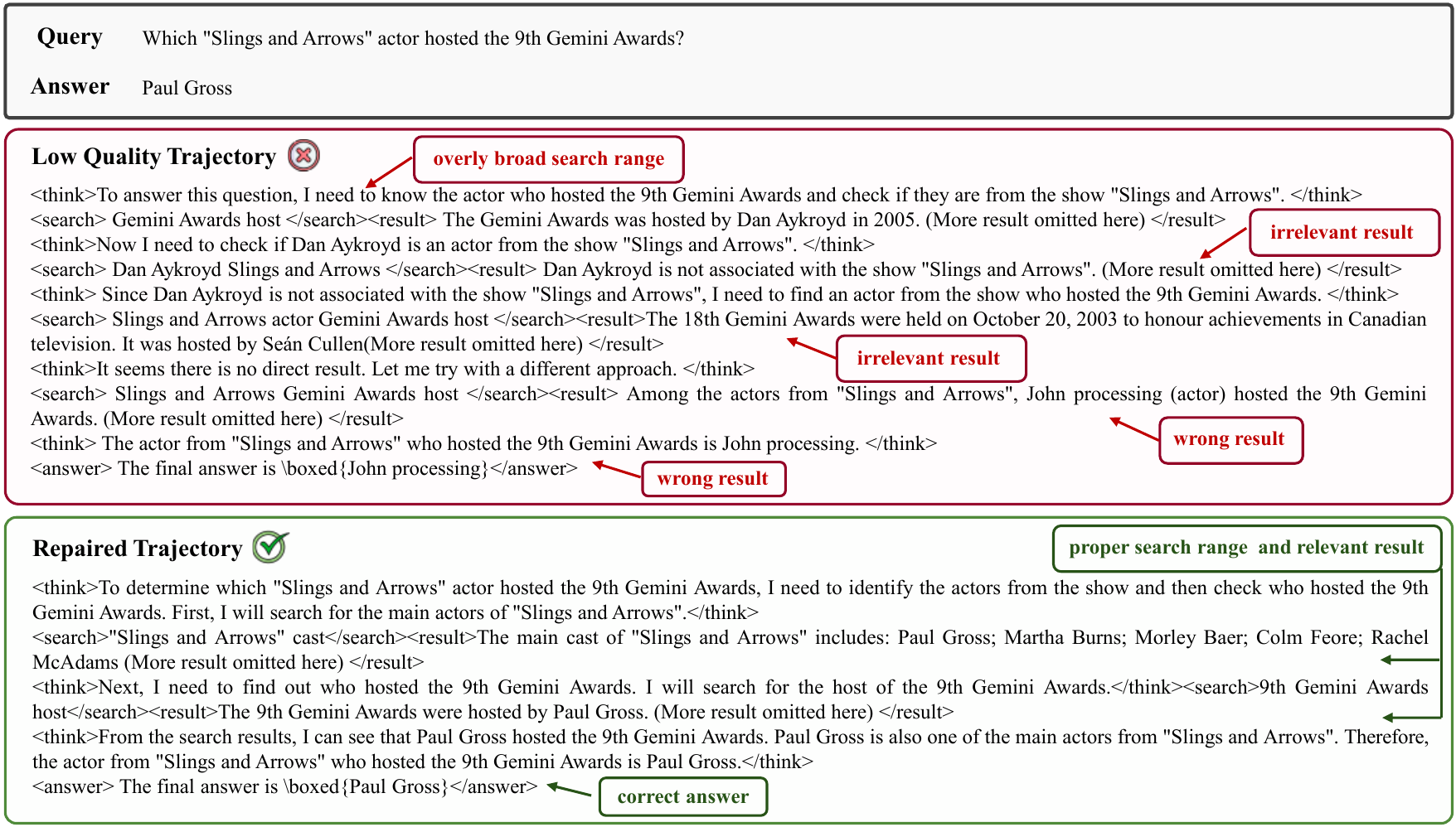}
 \caption{
  A Case study for an answer wrong low quality trajectory repair.
 }
 \label{repaircasestudy-2}

\end{figure*}

For an answer-wrong low-quality trajectory, as shown in Figure \ref{repaircasestudy-2}, the repairer first identifies that the low quality trajectory suffers from an overly broad search range at the initial stage. In addition, some search results return irrelevant information, which not only fails to assist the reasoning process but also interferes with it, ultimately leading to an incorrect final answer. In the repaired trajectory, this issue is effectively corrected: the repairer precisely targets the required search content, and the multiple search steps exhibit high relevance to one another. As a result, the model successfully reasons to the correct answer, thereby avoiding the issues present in the low-quality trajectory.

\subsection{Cases of AutoTraj}
\label{case study}
\begin{figure*}[t]
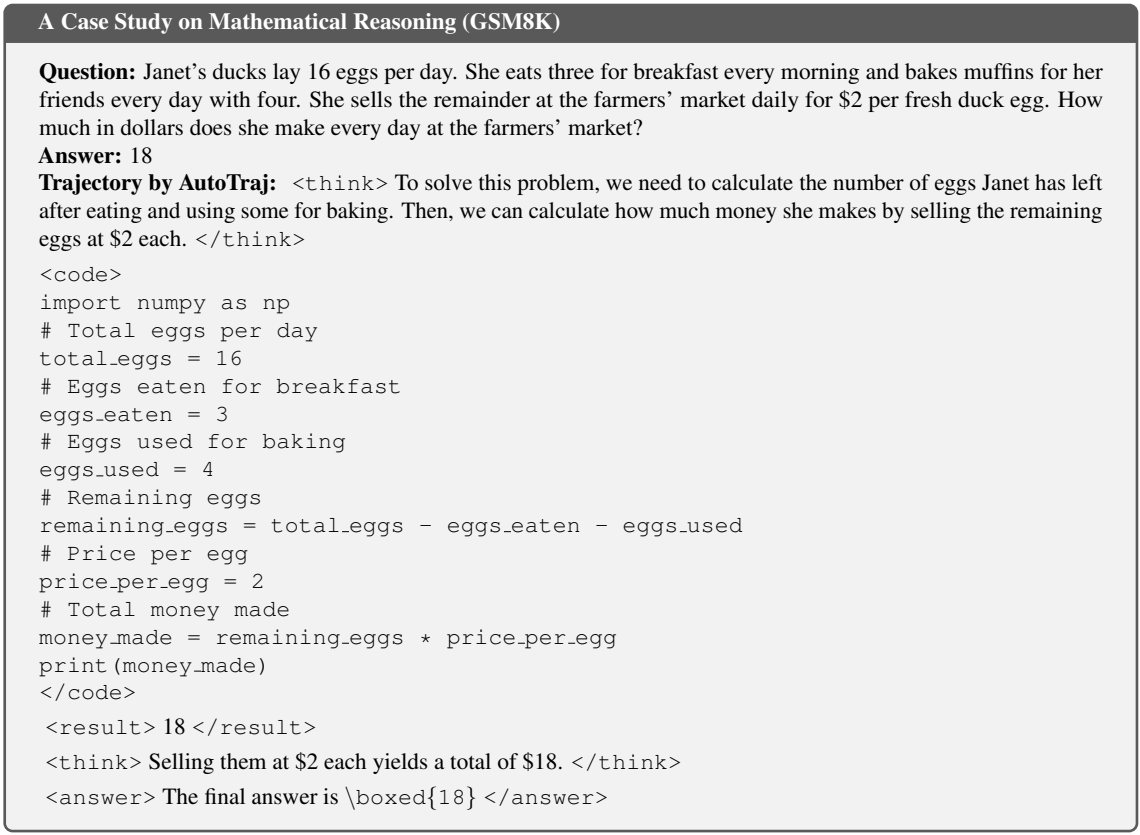

  \centering
  \scalebox{0.88}{
\begin{tcolorbox}[colframe=black!65!white, colback=black!50!white!10, coltitle=white, boxrule=0.4mm, arc=1mm, title=\textbf{A Case Study on Mathematical Reasoning (GSM8K)}]
  \textbf{Question:} \textnormal{Janet’s ducks lay 16 eggs per day. She eats three for breakfast every morning and bakes muffins for her friends every day with four. She sells the remainder at the farmers' market daily for \$2 per fresh duck egg. How much in dollars does she make every day at the farmers' market?}
  
  \textbf{Answer:} \textnormal{18}
  
  \textbf{Trajectory by AutoTraj:} \textnormal{
\texttt{<think>} To solve this problem, we need to calculate the number of eggs Janet has left after eating and using some for baking. Then, we can calculate how much money she makes by selling the remaining eggs at \$2 each. \texttt{</think>}
}

\vspace{0.3em}

\texttt{<code>} \\
\texttt{import numpy as np} \\
\texttt{\# Total eggs per day} \\
\texttt{total\_eggs = 16} \\
\texttt{\# Eggs eaten for breakfast} \\
\texttt{eggs\_eaten = 3} \\
\texttt{\# Eggs used for baking} \\
\texttt{eggs\_used = 4} \\
\texttt{\# Remaining eggs} \\
\texttt{remaining\_eggs = total\_eggs - eggs\_eaten - eggs\_used} \\
\texttt{\# Price per egg} \\
\texttt{price\_per\_egg = 2} \\
\texttt{\# Total money made} \\
\texttt{money\_made = remaining\_eggs * price\_per\_egg} \\
\texttt{print(money\_made)} \\
\texttt{</code>}

\vspace{0.3em}

\textnormal{
\texttt{<result>} 18 \texttt{</result>}
}

\vspace{0.3em}

\textnormal{
\texttt{<think>} Selling them at \$2 each yields a total of \$18. \texttt{</think>}
}

\vspace{0.3em}

\textnormal{
\texttt{<answer>} The final answer is \texttt{\textbackslash boxed\{18\}} \texttt{</answer>}
}
\end{tcolorbox}}
\caption{A Case Study on Mathematical Reasoning (GSM8K).}
\label{casestudy-1}
\end{figure*}

We further conduct case studies to illustrate how AutoTraj performs TIR in practice. As shown in Figure~\ref{casestudy-1} and Figure~\ref{casestudy-3}, when solving mathematical reasoning tasks, AutoTraj is able to autonomously select and invoke a code execution tool whenever necessary to assist the reasoning process and generate the final answer. As shown in Figure~\ref{casestudy-2} and Figure~\ref{casestudy-4}, when handling knowledge-intensive question answering tasks, AutoTraj dynamically chooses a search tool at appropriate stages of reasoning to retrieve relevant information, which subsequently guides the reasoning process toward producing the correct answer.

\begin{figure*}[t]
  \centering
  \scalebox{0.88}{
\begin{tcolorbox}[colframe=black!65!white, colback=black!50!white!10, coltitle=white, boxrule=0.4mm, arc=1mm, title=\textbf{A Case Study on Mathematical Reasoning (Math)}]
  \textbf{Question:} \textnormal{At 2:30 p.m. during a long drive, Bobbi asks her parents, ``Are we there yet?" Her mother responds, ``We will be there in 7200 seconds." If Bobbi's mother is correct, at what time in the afternoon will they arrive at their destination?}
  
  \textbf{Answer:} \textnormal{4:30 p.m.}
  
  \textbf{Trajectory by AutoTraj:} \textnormal{
\texttt{\detokenize{<think>}} \\
To solve this problem, I need to convert 7200 seconds into hours and then add that to the current time of 2:30 p.m. \\
\texttt{\detokenize{</think>}} \\
\texttt{\detokenize{<code>}} \\
\texttt{import datetime} \\
\texttt{\# Current time} \\
\texttt{current\_time = datetime.datetime.strptime('14:30', '\%H:\%M')} \\
\texttt{\# Time in seconds} \\
\texttt{time\_in\_seconds = 7200} \\
\texttt{\# Convert seconds to timedelta} \\
\texttt{time\_delta = datetime.timedelta(seconds=time\_in\_seconds)} \\
\texttt{\# Calculate the arrival time} \\
\texttt{arrival\_time = current\_time + time\_delta} \\
\texttt{\# Format the arrival time} \\
\texttt{arrival\_time\_format = arrival\_time.strftime('\%I:\%M \%p')} \\
\texttt{print(arrival\_time\_format)} \\
\texttt{\detokenize{</code>}} \\
\texttt{\detokenize{<result>}} 
04:30 PM 
\texttt{\detokenize{</result>}} \\
\texttt{\detokenize{<think>}} 
The arrival time is 4:30 p.m. Therefore, Bobbi and her parents will arrive at their destination at 4:30 p.m. 
\texttt{\detokenize{</think>}} \\ 
\texttt{\detokenize{<answer>}} 
The final answer is \texttt{\textbackslash boxed\{4:30 p.m.\}} 
\texttt{\detokenize{</answer>}}
}
\end{tcolorbox}}
\caption{A Case Study on Mathematical Reasoning (Math).}
\label{casestudy-3}
\end{figure*}

\begin{figure*}[t]
  \centering
  \scalebox{0.88}{
\begin{tcolorbox}[colframe=black!65!white, colback=black!50!white!10, coltitle=white, boxrule=0.4mm, arc=1mm, title=\textbf{A Case Study on Knowledge-Intensive Reasoning (HotpotQA)}]
  \textbf{Question:} \textnormal{Which of these battles occurred earlier--the Battle of Peleliu or the Battle of Belleau Wood?}
  
  \textbf{Answer:} \textnormal{The Battle of Belleau Wood}
  
  \textbf{Trajectory by AutoTraj:} \textnormal{
\texttt{\detokenize{<think>}}
To answer this question, I need to compare the dates of the Battle of Peleliu and the Battle of Belleau Wood.
\texttt{\detokenize{</think>}}
\texttt{\detokenize{<search>}} Battle of Peleliu date \texttt{\detokenize{</search>}}
\texttt{\detokenize{<result>}}
The Battle of Peleliu, codenamed Operation Stalemate II by the United States military, was fought between the U.S. and Japan during the Mariana and Palau Campaign... (More result omitted here)
\texttt{\detokenize{</result>}}
\texttt{\detokenize{<think>}}
Now I know the date of the Battle of Peleliu. Next, I will search for the date of the Battle of Belleau Wood.
\texttt{\detokenize{</think>}}
\texttt{\detokenize{<search>}} Battle of Belleau Wood date \texttt{\detokenize{</search>}}
\texttt{\detokenize{<result>}} 
More result omitted here
\texttt{\detokenize{</result>}}
\texttt{\detokenize{<think>}}
Now I have the dates for both battles. The Battle of Peleliu took place from September 15, 1944, to November 25, 1944, and the Battle of Belleau Wood took place from June 1, 1916, to June 26, 1916. Therefore, the Battle of Belleau Wood occurred earlier than the Battle of Peleliu.
\texttt{\detokenize{</think>}}
\texttt{\detokenize{<answer>}}
The Battle of Belleau Wood occurred earlier than the Battle of Peleliu. Therefore, the answer is \texttt{\detokenize{\boxed{The Battle of Belleau Wood}}}
\texttt{\detokenize{</answer>}}
}
\end{tcolorbox}}
\caption{A Case Study on Knowledge-Intensive Reasoning (HotpotQA).}
\label{casestudy-2}
\end{figure*}

\begin{figure*}[t]
  \centering
  \scalebox{0.88}{
\begin{tcolorbox}[colframe=black!65!white, colback=black!50!white!10, coltitle=white, boxrule=0.4mm, arc=1mm, title=\textbf{A Case Study on Knowledge-Intensive Reasoning (2WikiMultiHopQA)}]
  \textbf{Question:} \textnormal{What is the place of birth of Gwilym Lloyd George's father?}
  
  \textbf{Answer:} \textnormal{Manchester}
  
  \textbf{Trajectory by AutoTraj:} \textnormal{
\texttt{\detokenize{<think>}} 
To answer this question, I need to know the place of birth of Gwilym Lloyd George's father. 
\texttt{\detokenize{</think>}} 
\texttt{\detokenize{<search>}} Gwilym Lloyd George \texttt{\detokenize{</search>}} 
\texttt{\detokenize{<result>}} 
Gwilym Lloyd George, 1st Viscount Tenby, (4 December 1894 -- 14 February 1967) was a British politician and cabinet minister. A younger son of Prime Minister David Lloyd George... (More result omitted here)
\texttt{\detokenize{</result>}} 
\texttt{\detokenize{<think>}} 
Now that I know Gwilym Lloyd George's father is David Lloyd George, I need to find the place of his birth. 
\texttt{\detokenize{</think>}} 
\texttt{\detokenize{<search>}} David Lloyd George place of birth \texttt{\detokenize{</search>}} 
\texttt{\detokenize{<result>}} 
(More result omitted here)
\texttt{\detokenize{</result>}} 
\texttt{\detokenize{<think>}} 
The place of birth of David Lloyd George, Gwilym Lloyd George's father, is Manchester, England. 
\texttt{\detokenize{</think>}} 
\texttt{\detokenize{<answer>}} 
The place of birth of Gwilym Lloyd George's father is Manchester. Therefore, the answer is \texttt{\detokenize{\boxed{Manchester}}} 
\texttt{\detokenize{</answer>}}
}
\end{tcolorbox}}
\caption{A Case Study on Knowledge-Intensive Reasoning (2WikiMultiHopQA).}
\label{casestudy-4}
\end{figure*}

\end{document}